%% file: emnlp2020.tex
\documentclass[11pt,a4paper]{article}
\usepackage[hyperref]{emnlp2020}
\usepackage{times}
\usepackage{graphicx}
\usepackage{latexsym}

\usepackage{multirow}
\usepackage{url}
\usepackage{amssymb}
\usepackage{amsmath}
\usepackage{enumitem}
\usepackage{color,soul}
\usepackage{subfig}

\usepackage{microtype}

\aclfinalcopy 
\def\aclpaperid{1201} 

\title{TOD-BERT: Pre-trained Natural Language Understanding for Task-Oriented Dialogue}

\author{Chien-Sheng Wu, Steven Hoi, Richard Socher, and Caiming Xiong \\
Salesforce Research \\
\texttt{[cswu, shoi, rsocher, cxiong]@salesforce.com}
}

\date{}

\begin{document}
\maketitle
\begin{abstract}
The underlying difference of linguistic patterns between general text and task-oriented dialogue makes existing pre-trained language models less useful in practice.
In this work, we unify nine human-human and multi-turn task-oriented dialogue datasets for language modeling. 
To better model dialogue behavior during pre-training, we incorporate user and system tokens into the masked language modeling. We propose a contrastive objective function to simulate the response selection task.
Our pre-trained task-oriented dialogue BERT (TOD-BERT) outperforms strong baselines like BERT on four downstream task-oriented dialogue applications, including intention recognition, dialogue state tracking, dialogue act prediction, and response selection.
We also show that TOD-BERT has a stronger few-shot ability that can mitigate the data scarcity problem for task-oriented dialogue.
\end{abstract}

\section{Introduction}
\label{sec:Introduction}

Pre-trained models with self-attention encoder architectures \cite{devlin2018bert, liu2019roberta} have been commonly used in many NLP applications. 
Such models are self-supervised based on a massive scale of general text corpora, such as English Wikipedia or books \cite{zhu2015aligning}. 
By further fine-tuning these representations, breakthroughs have been continuously reported for various downstream tasks, especially natural language understanding.

However, previous work \cite{rashkin2018towards, wolf2019transfertransfo} shows that there are some deficiencies in the performance to apply fine-tuning on conversational corpora directly.
One possible reason could be the intrinsic difference of linguistic patterns between human conversations and writing text, resulting in a large gap of data distributions \cite{bao2019plato}. Therefore, pre-training dialogue language models using chit-chat corpora from social media, such as Twitter or Reddit, has been recently investigated, especially for dialogue response generation \cite{zhang2019dialogpt} and retrieval \cite{henderson-etal-2019-training}. Although these open-domain dialogues are diverse and easy-to-get, they are usually short, noisy, and without specific chatting goals.

On the other hand, a task-oriented dialogue has explicit goals (e.g. restaurant reservation or ticket booking) and many conversational interactions. But each dataset is usually small and scattered because obtaining and labeling such data is time-consuming. Moreover, a task-oriented dialogue has explicit user and system behaviors where a user has his/her goal, and a system has its belief and database information, which makes the language understanding component and dialogue policy learning more important than those chit-chat scenarios. 

This paper aims to prove this hypothesis: self-supervised language model pre-training using task-oriented corpora can learn better representations than existing pre-trained models for task-oriented downstream tasks. We emphasize that what we care about the most is not whether our pre-trained model can achieve state-of-the-art results on each downstream task since most of the current best models are built on top of pre-trained models, and ours can easily replace them. We avoid adding too many additional components on top of the pre-training architecture when fine-tuning in our experiments.

We collect and combine nine human-human and multi-turn task-oriented dialogue corpora to train a task-oriented dialogue BERT (TOD-BERT). In total, there are around 100k dialogues with 1.4M utterances across over 60 different domains. Like BERT~\cite{devlin2018bert}, TOD-BERT is formulated as a masked language model and uses a deep bidirectional Transformer \cite{vaswani2017attention} encoder as its model architecture. Unlike BERT, TOD-BERT incorporates two special tokens for user and system to model the corresponding dialogue behavior. A contrastive objective function of response selection task is combined during pre-training stage to capture response similarity. We select BERT because it is the most widely used model in NLP research recently, and our unified datasets can be easily applied to pre-train any existing language models.

We test TOD-BERT on task-oriented dialogue systems on four core downstream tasks, including intention recognition, dialogue state tracking, dialogue act prediction, and response selection. What we observe is: TOD-BERT outperforms BERT and other strong baselines such as GPT-2~\cite{radford2019language} and DialoGPT~\cite{zhang2019dialogpt} on all the selected downstream tasks, which further confirms its effectiveness for improving dialogue language understanding. We find that response contrastive learning is beneficial, but it is currently overlooked not well-investigated in dialogue pre-training research. More importantly, TOD-BERT has a stronger few-shot ability than BERT on each task, suggesting that it can reduce the need for expensive human-annotated labels. TOD-BERT can be easily leveraged and adapted to a new task-oriented dialogue dataset.
Our source code and data processing are released to facilitate future research on pre-training and fine-tuning of task-oriented dialogue~\footnote{\url{github.com/jasonwu0731/ToD-BERT}}.

\section{Related Work}
\label{sec:related_work}

\paragraph{General Pre-trained Language Models,} which are trained on massive general text such as Wikipedia and BookCorpus, can be roughly divided into two categories: uni-directional or bi-directional attention mechanisms. GPT \cite{radford2018improving} and GPT-2 \cite{radford2019language} are representatives of uni-directional language models using a Transformer decoder, where the objective is to maximize left-to-right generation likelihood. These models are commonly applied in natural language generation tasks. On the other hand, BERT \cite{devlin2018bert}, RoBERTa \cite{liu2019roberta}, and their variances are pre-trained using a Transformer encoder with bi-directional token prediction. These models are usually evaluated on classification tasks such as GLUE benchmark \cite{wang2018glue} or span-based question answering tasks \cite{rajpurkar2016squad}. 

Some language models can support both uni-directional and bi-directional attention, such as UniLM \cite{dong2019unified}.
Conditional language model pre-training is also proposed. For example, CTRL \cite{keskar2019ctrl} is a conditional Transformer model, trained to condition on control codes that govern style, content, and task-specific behavior.
Recently, multi-task language model pre-training with unified sequence-to-sequence generation is proposed. Text-to-text Transformer (T5) \cite{raffel2019exploring} unifies multiple text modeling tasks and achieves the promising results in various NLP benchmarks.

\paragraph{Dialogue Pre-trained Language Models} are mostly trained on open-domain conversational data from Reddit or Twitter for dialogue response generation. 
Transfertransfo \cite{wolf2019transfertransfo} achieves good performance on ConvAI-2 dialogue competition using GPT-2.
DialoGPT \cite{zhang2019dialogpt} is an extension of GPT-2 that is pre-trained on Reddit data for open-domain response generation.
ConveRT \cite{henderson2019convert} pre-trained a dual transformer encoder for response selection task on large-scale Reddit (input, response) pairs. 
PLATO \cite{bao2019plato} uses both Twitter and Reddit data to pre-trained a dialogue generation model with discrete latent variables.
All of them are designed to cope with the response generation task for open-domain chatbots.

\begin{table*}
\centering
\resizebox{0.95\linewidth}{!}{
\begin{tabular}{r|c|c|c|c}
\hline
\textbf{Name} & \textbf{\# Dialogue} & \textbf{\# Utterance} & \textbf{Avg. Turn} & \textbf{\# Domain} \\ \hline
MetaLWOZ~\cite{lee2019multi-domain} & 37,884 & 432,036 & 11.4 & 47 \\ \hline
Schema~\cite{rastogi2019towards} & 22,825 & 463,284 & 20.3 & 17 \\ \hline
Taskmaster~\cite{byrne2019taskmaster} & 13,215 & 303,066 & 22.9 & 6 \\ \hline
MWOZ~\cite{budzianowski2018multiwoz} & 10,420 & 71,410 & 6.9 & 7 \\ \hline
MSR-E2E~\cite{li2018microsoft} & 10,087 & 74,686 & 7.4 & 3 \\ \hline
SMD~\cite{eric2017key} & 3,031 & 15,928 & 5.3 & 3 \\ \hline
Frames~\cite{asri2017frames} & 1,369 & 19,986 & 14.6 & 3 \\ \hline
WOZ~\cite{mrkvsic2016neural} & 1,200 & 5,012 & 4.2 & 1 \\ \hline
CamRest676~\cite{wen2016network} & 676 & 2,744 & 4.1 & 1 \\ \hline
\end{tabular}
}
\caption{Data statistics for task-oriented dialogue datasets.}
\label{tb:train_dataset}
\end{table*}

Pretraining for task-oriented dialogues, on the other hand, has few related works. \citet{budzianowski2019hello} first apply the GPT-2 model to train on response generation task, which takes system belief, database result, and last dialogue turn as input to predict next system responses. It only uses one dataset to train its model because few public datasets have database information available. \citet{henderson-etal-2019-training} pre-trained a response selection model for task-oriented dialogues. They first pre-train on Reddit corpora and then fine-tune on target dialogue domains, but their training and fine-tuning code is not released. \citet{peng2020few} focus on the natural language generation (NLG) task, which assumes dialogue acts and slot-tagging results are given to generate a natural language response. Pre-training on a set of annotated NLG corpora can improve conditional generation quality using a GPT-2 model.

\section{Method}
\label{sec:Method}
This section discusses each dataset used in our task-oriented pre-training and how we process the data. Then we introduce the selected pre-training base model and its objective functions.

\subsection{Datasets}
\label{subsec:Datasets}
We collect nine different task-oriented datasets which are English, human-human and multi-turn. In total, there are 100,707 dialogues, which contain 1,388,152 utterances over 60 domains. Dataset statistics is shown in Table~\ref{tb:train_dataset}.

\begin{itemize}[leftmargin=*]
\item \textbf{MetaLWOZ}~\cite{lee2019multi-domain}: 
Meta-Learning Wizard-of-Oz is a dataset designed to help develop models capable of predicting user responses in unseen domains. This large dataset was created by crowdsourcing 37,884 goal-oriented dialogs, covering 227 tasks in 47 domains. The MetaLWOZ dataset is used as the fast adaptation task for DSTC8 \cite{DSTC8} dialogue competition.

\item \textbf{Schema}~\cite{rastogi2019towards}: Schema-guided dialogue has 22,825 dialogues and provides a challenging testbed for several tasks, in particular, dialogue state tracking. Each schema is a set of tracking slots, and each domain could have multiple possible schemas. This allows a single dialogue system to support many services and facilitates the simple integration of new services without requiring much training data. The Schema dataset is used as the dialogue state tracking task for DSTC8 \cite{DSTC8} dialogue competition.

\item \textbf{Taskmaster}~\cite{byrne2019taskmaster}:
This dataset includes 13,215 dialogues comprising six domains, including 5,507 spoken and 7,708 written dialogs created with two distinct procedures. One is a two-person Wizard of Oz approach that one person acts like a robot, and the other is a self-dialogue approach in which crowdsourced workers wrote the entire dialog themselves. It has 22.9 average conversational turns in a single dialogue, which is the longest among all task-oriented datasets listed.

\item \textbf{MWOZ}~\cite{budzianowski2018multiwoz}:
Multi-Domain Wizard-of-Oz dataset contains 10,420 dialogues over seven domains, and it has multiple domains in a single dialogue. It has a detailed description of the data collection procedure, user goal, system act, and dialogue state labels. Different from most of the existing corpora, it also provides full database information. 

\item \textbf{MSR-E2E}~\cite{li2018microsoft}:
Microsoft end-to-end dialogue challenge has 10,087 dialogues in three domains, movie-ticket booking, restaurant reservation, and taxi booking. It also includes an experiment platform with built-in simulators in each
domain. 

\item \textbf{SMD}~\cite{eric2017key}:
Stanford multi-domain dialogue is an in-car personal assistant dataset, comprising 3,301 dialogues and three domains: calendar scheduling, weather information retrieval, and point-of-interest navigation. It is designed to smoothly interface with knowledge bases, where a knowledge snippet is attached with each dialogue as a piece of simplified database information. 

\item \textbf{Frames}~\cite{asri2017frames}:
This dataset comprises 1,369 human-human dialogues with an average of 14.6 turns per dialogue, where users are given some constraints to book a trip and assistants who search a database to find appropriate trips. Unlike other datasets, it has labels to keep track of different semantic frames, which is the decision-making behavior of users throughout each dialogue.

\item \textbf{WOZ}~\cite{mrkvsic2016neural} and \textbf{CamRest676}~\cite{wen2016network}:
These two corpora use the same data collection procedure and same ontology from DSTC2 \cite{henderson-etal-2014-second}. They are one of the first task-oriented dialogue datasets that use Wizard of Oz style with text input instead of speech input, which improves the model’s capacity for the semantic understanding instead of its robustness to automatic speech recognition errors.

\end{itemize}



\subsection{TOD-BERT}
\label{subsec:Model}
We train our TOD-BERT based on BERT architecture using two loss functions: masked language modeling (MLM) loss and response contrastive loss (RCL). Note that the datasets we used can be used to pre-train any existing language model architecture, and here we select BERT because it is the most widely used model in NLP research. We use the BERT-base uncased model, which is a transformer self-attention encoder \cite{vaswani2017attention} with 12 layers and 12 attention heads with its hidden size $d_{B} = 768$. 

\begin{figure}[t]
\centering
\includegraphics[width=\linewidth]{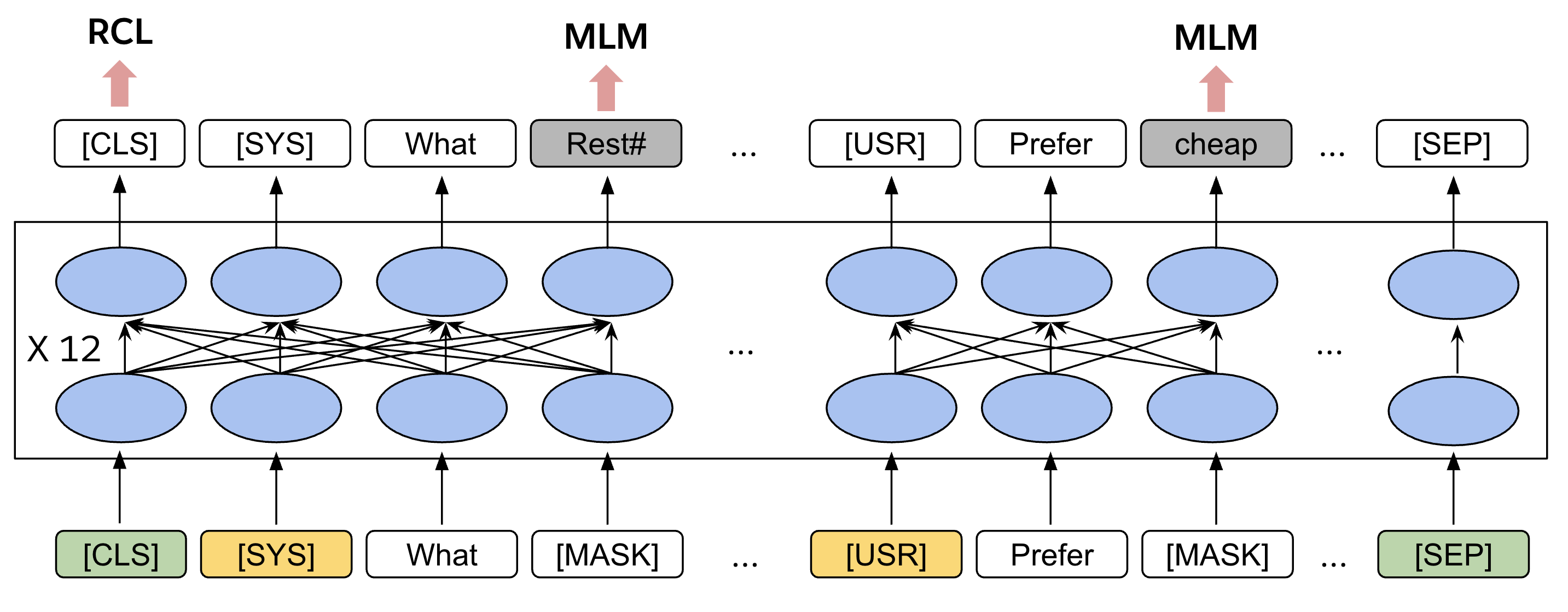}
\caption{Dialogue pre-training based on Transformer encoder with user and system special tokens. Two objective functions are used: masked language modeling and response contrastive learning.}
\label{fig:model}
\end{figure}

To capture speaker information and the underlying interaction behavior in dialogue, we add two special tokens, [USR] and [SYS], to the byte-pair embeddings \cite{mrkvsic2016neural}. We prefix the special token to each user utterance and system response, and concatenate all the utterances in the same dialogue into one flat sequence, as shown in Figure~\ref{fig:model}. For example, for a dialogue $D=\{S_1, U_1,\dots, S_n, U_n\}$, where $n$ is the number of dialogue turns and each $S_i$ or $U_i$ contains a sequence of words, the input of the pre-training model is processed as ``[SYS] $S_1$ [USR] $U_1 \dots$ '' with standard positional embeddings and segmentation embeddings.

\paragraph{Masked language modeling} is a common pre-training strategy for BERT-like architectures, in which a random sample of tokens in the input sequence is selected and replaced with the special token [MASK]. The MLM loss function is the cross-entropy loss on predicting the masked tokens. In the original implementation, random masking and replacement are performed once in the beginning and saved for the training duration. Here we conduct token masking dynamically during batch training. TOD-BERT is initialized from BERT, a good starting parameter set, then is further pre-trained on those task-oriented corpora. The MLM loss function is
\begin{equation}
\begin{array}{c}
    L_{mlm} = - \sum_{m=1}^{M} \log P(x_m),
\end{array}
\end{equation}
where $M$ is the total number of masked tokens and $P(x_m)$ is the predicted probability of the token $x_m$ over the vocabulary size.

\paragraph{Response contrastive loss} can also be used for dialogue language modeling since it does not require any additional human annotation. Pre-training with RCL can bring us several advantages: 1) we can learn a better representation for the [CLS] token, as it is essential for all the downstream tasks, and 2) we encourage the model to capture underlying dialogue sequential order, structure information, and response similarity. 

Unlike the original next sentence prediction (NSP) objective in BERT pre-training, which concatenates two segments $A$ and $B$ to predict whether they are consecutive text with binary classification, we apply a dual-encoder approach~\cite{henderson2019convert} and simulate multiple negative samples. We first draw a batch of dialogues $\{D_1, \dots, D_b\}$ and split each dialogue at a randomly selected turn $t$. For example, $D_1$ will be separated into two segments, one is the context $\{S^1_1, U^1_1, \dots, S^1_t, U^1_t\}$ and the other is the response $\{S^1_{t+1}\}$. We use TOD-BERT to encode all the contexts and their corresponding responses separately.

Afterwards, we have a context matrix $C \in \mathbb{R}^{b \times d_B}$ and a response matrix $R \in \mathbb{R}^{b \times d_B}$ by taking the output [CLS] representations from the $b$ dialogues. We treat other responses in the same batch as randomly selected negative samples. The RCL objective function is
\begin{equation}
\begin{array}{c}
    L_{rcl} = - \sum\limits_{i=1}^{b} \log M_{i, i}, \\ 
    M = \textnormal{Softmax}(C R^T) \in \mathbb{R}^{b \times b}. 
\end{array}
\end{equation}
Increasing batch size to a certain amount can obtain better performance on downstream tasks, especially for the response selection. The Softmax function normalizes the vector per row. In our setting, increasing batch size also means changing the positive and negative ratio in the contrastive learning. Batch size is a hyper-parameter that may be limited by hardware. We also try different negative sampling strategies during pre-training such as local sampling~\cite{saeidi2017effect}, but do not observe significant change compared to random sampling.

\paragraph{Overall} pre-training loss function is the weighted-sum of $L_{mlm}$ and $L_{rcl}$, and in our experiments, we simply sum them up. We gradually reduce the learning rate without a warm-up period. We train TOD-BERT with AdamW \cite{loshchilov2017decoupled} optimizer with a dropout ratio of 0.1 on all layers and attention weights. GELU activation functions \cite{hendrycks2016gaussian} is used. Models are early-stopped using perplexity scores of a held-out development set, with mini-batches containing 32 sequences of maximum length 512 tokens. Experiments are conducted on two NVIDIA Tesla V100 GPUs.

\section{Downstream Tasks}
We care the most in this paper whether TOD-BERT, a pre-trained language model using aggregated task-oriented corpora, can show any advantage over BERT. Therefore, we avoid adding too many additional components on top of its architecture when fine-tuning on each downstream task. Also, we use the same architecture with a similar number of parameters for a fair comparison. All the model parameters are updated with a gradient clipping to 1.0 using the same hyper-parameters during fine-tuning. We select four crucial task-oriented downstream tasks to evaluate: intent recognition, dialogue state tracking, dialogue act prediction, and response selection. All of them are core components in modularized task-oriented systems~\cite{wen2016network}. 

\paragraph{Intent recognition} task is a multi-class classification problem, where we input a sentence $U$ and models predict one single intent class over $I$ possible intents.  
\begin{equation}
\begin{array}{c}
    P_{int} = \textnormal{Softmax}(W_1(F(U))) \in \mathbb{R}^{I},
\end{array}
\end{equation}
where $F$ is a pre-trained language model and we use its [CLS] embeddings as the output representation. $W_1 \in \mathbb{R}^{I \times d_{B}}$ is a trainable linear mapping. The model is trained with cross-entropy loss between the predicted distributions $P_{int}$ and the true intent labels.

\paragraph{Dialogue state tracking} can be treated as a multi-class classification problem using a predefined ontology. Unlike intent, we use dialogue history $X$ (a sequence of utterances) as input and a model predicts slot values for each (domain, slot) pair at each dialogue turn. Each corresponding value $v^{j}_i$, the $i$-th value for the $j$-th (domain, slot) pair, is passed into a pre-trained model and fixed its representation during training. 
\begin{equation}
\begin{array}{c}
    S^{j}_i = Sim(G_{j}(F(X)), F(v^{j}_i)) \in \mathbb{R}^{1}, \\
\end{array}
\end{equation}
where $Sim$ is the cosine similarity function, and $S^{j} \in \mathbb{R}^{|v^j|}$ is the probability distribution of the $j$-th (domain, slot) pair over its possible values. $G_{j}$ is the slot projection layer of the $j$ slot, and the number of layers $|G|$ is equal to the number of (domain, slot) pairs. The model is trained with cross-entropy loss summed over all the pairs.

\paragraph{Dialogue act prediction} is a multi-label classification problem because a system response may contain multiple dialogue acts, e.g., request and inform at the same time. Model take dialogue history as input and predict a binary result for each possible dialogue act:
\begin{equation}
\begin{array}{c}
    A = Sigmoid(W_2(F(X))) \in \mathbb{R}^{N}, 
\end{array}
\end{equation}
where $W_2 \in \mathbb{R}^{d_{B} \times N}$ is a trainable linear mapping, $N$ is the number of possible dialogue acts, and each value in $A$ is between $[0, 1]$ after a Sigmoid layer. The model is trained with binary cross-entropy loss and the $i$-th dialogue act is considered as a triggered dialogue act if $A_i > 0.5$.

\paragraph{Response selection} is a ranking problem, aiming to retrieve the most relative system response from a candidate pool. We use a dual-encoder strategy \cite{henderson-etal-2019-training} and compute similarity scores between source $X$ and target $Y$, 
\begin{equation}
\begin{array}{c}
    r_{i} = Sim(F(X), F(Y_i)) \in \mathbb{R}^{1},
\end{array}
\end{equation}
where $Y_i$ is the $i$-th response candidate and $r_i$ is its cosine similarity score. Source $X$ can be truncated, and we limit the context lengths to the most recent 256 tokens in our experiments. We randomly sample several system responses from the corpus as negative samples. Although it may not be a true negative sample, it is common to train a ranker and evaluate its results \cite{henderson2019convert}.

\section{Evaluation Datasets}
We pick up several datasets, OOS, DSTC2, GSIM, and MWOZ, for downstream evaluation. The first three corpora are not included in the pre-trained task-oriented datasets. For MWOZ, to be fair, we do not include its test set dialogues during the pre-training stage. Details of each evaluation dataset are discussed in the following:

\begin{itemize}[leftmargin=*]

\item \textbf{OOS}~\cite{larson2019evaluation}: 
The out-of-scope intent dataset is one of the largest annotated intent datasets, including 15,100/3,100/5,500 samples for the train, validation, and test sets, respectively. It covers 151 intent classes over ten domains, including 150 in-scope intent and one out-of-scope intent. The out-of-scope intent means that a user utterance that does not fall into any of the predefined intents. Each of the intents has 100 training samples. 

\item \textbf{DSTC2}~\cite{henderson-etal-2014-second}: 
DSTC2 is a human-machine task-oriented dataset that may include a certain system response noise. It has 1,612/506/1117 dialogues for train, validation, and test sets, respectively. We follow \citet{paul2019towards} to map the original dialogue act labels to universal dialogue acts, which results in 19 different system dialogue acts. 

\item \textbf{GSIM}~\cite{shah2018bootstrapping}:
GSIM is a human-rewrote machine-machine task-oriented corpus, including 1500/469/1039 dialogues for the train, validation, and test sets, respectively. We combine its two domains, movie and restaurant domains, into one single corpus. It is collected by Machines
Talking To Machines (M2M) \cite{shah2018building} approach, a functionality-driven process combining a dialogue self-play step and a crowdsourcing step. We map its dialogue act labels to universal dialogue acts \cite{paul2019towards}, resulting in 13 different system dialogue acts. 

\item \textbf{MWOZ}~\cite{budzianowski2018multiwoz}: MWOZ is the most common benchmark for task-oriented dialogues, especially for dialogue state tracking. It has 8420/1000/1000 dialogues for train, validation, and test sets, respectively. Across seven different domains, in total, it has 30 (domain, slot) pairs that need to be tracked in the test set. We use its revised version MWOZ 2.1, which has the same dialogue transcripts but with cleaner state label annotation. 
\end{itemize}

\begin{table*}[t]
\centering
\resizebox{0.8\linewidth}{!}{

\begin{tabular}{r|r|c|c|c|c}
\hline
\multicolumn{1}{r|}{} & \multicolumn{1}{c|}{\textbf{Model}} & \textbf{\begin{tabular}[c]{@{}c@{}}Acc\\ (all)\end{tabular}} & \textbf{\begin{tabular}[c]{@{}c@{}}Acc\\ (in)\end{tabular}} & \textbf{\begin{tabular}[c]{@{}c@{}}Acc\\ (out)\end{tabular}} & \textbf{\begin{tabular}[c]{@{}c@{}}Recall\\ (out)\end{tabular}} \\ \hline
\multirow{2}{*}{\textbf{1-Shot}} 
 & BERT & 29.3\% $\pm$ 3.4\% & 35.7\% $\pm$ 4.1\% & 81.3\% $\pm$ 0.4\% & 0.4\% $\pm$ 0.3\% \\
 & TOD-BERT-mlm & 38.9\% $\pm$ 6.3\% & 47.4\% $\pm$ 7.6\% & 81.6\% $\pm$ 0.2\% & \textbf{0.5\%} $\pm$ 0.2\% \\
 & TOD-BERT-jnt & \textbf{42.5\%} $\pm$ 0.1\% & \textbf{52.0\%} $\pm$ 0.1\% & \textbf{81.7\%} $\pm$ 0.1\% & 0.1\% $\pm$ 0.1\% \\
 \hline
\multirow{2}{*}{\textbf{10-Shot}} 
 & BERT & 75.5\% $\pm$ 1.1\% & 88.6\% $\pm$ 1.1\% & 84.7\% $\pm$ 0.3\%  & 16.5\% $\pm$ 1.7\% \\
 & TOD-BERT-mlm & 76.6\% $\pm$ 0.8\% & 90.5\% $\pm$ 1.2\% & 84.3\% $\pm$ 0.2\% & 14.0\% $\pm$ 1.3\% \\
 & TOD-BERT-jnt & \textbf{77.3\%} $\pm$ 0.5\% & \textbf{91.0\%} $\pm$ 0.5\% & \textbf{84.5\%} $\pm$ 0.4\% & \textbf{15.3\%} $\pm$ 2.1\% \\ 
 \hline
\multirow{8}{*}{\begin{tabular}[c]{@{}r@{}}\textbf{Full}\\ \textbf{(100-Shot)}\end{tabular}} 
 & FastText* & - & 89.0\% & - & 9.7\%  \\
 & SVM* & - & 91.0\% & - & 14.5\% \\ 
 & CNN* & - & 91.2\% & - & 18.9\% \\
 & GPT2 & 83.0\% & 94.1\% & 87.7\% & 32.0\% \\ 
 & DialoGPT & 83.9\% & 95.5\% & 87.6\% & 32.1\% \\
 & BERT & 84.9\% & 95.8\% & 88.1\% & 35.6\% \\  
 & TOD-BERT-mlm & 85.9\% & 96.1\% & 89.5\% & \textbf{46.3\%} \\
 & TOD-BERT-jnt  & \textbf{86.6\%} & \textbf{96.2\%} & \textbf{89.9\%} & 43.6\% \\
 \hline
\end{tabular}
}
\caption{Intent recognition results on the OOS dataset, one of the largest intent corpus. Models with * are reported from \citet{larson2019evaluation}.}
\label{tb:intent}
\end{table*}

\begin{table}[t]
\centering
\resizebox{\linewidth}{!}{
\begin{tabular}{r|c|c|c}
\hline
 & \begin{tabular}[c]{@{}c@{}}\textbf{Domain}\\ (acc)\end{tabular} & \begin{tabular}[c]{@{}c@{}}\textbf{Intent}\\ (acc)\end{tabular} &  \begin{tabular}[c]{@{}c@{}}\textbf{Dialogue Act}\\ (F1-micro)\end{tabular} \\ \hline
GPT2 & 63.5\% & 74.7\%  & 85.7\% \\
DialoGPT & 63.0\% & 65.7\% & 84.2\% \\
BERT & 60.5\% & 71.1\% &85.3\% \\
TOD-BERT-mlm & 63.9\% & 70.7\% & 83.5\% \\
TOD-BERT-jnt & \textbf{68.7\%} & \textbf{77.8\%} & \textbf{86.2\%} \\ \hline
\end{tabular}
}
\caption{Probing results of different pre-trained language models using a single-layer perceptron.}
\label{tb:probe}
\end{table}

\section{Results}
\label{sec:Results}
For each downstream task, we first conduct the experiments using the whole dataset, and then we simulate the few-shot setting to show the strength of our TOD-BERT. We run at least three times with different random seeds for each few-shot experiment to reduce data sampling variance, and we report its mean and standard deviation for these limited data scenarios. We investigate two versions of TOD-BERT; one is TOD-BERT-mlm that only uses MLM loss during pre-training, and the other is TOD-BERT-jnt, which is jointly trained with the MLM and RCL objectives. We compare TOD-BERT with BERT and other baselines, including two other strong pre-training models GPT-2~\cite{radford2019language} and DialoGPT~\cite{zhang2019dialogpt}. For a GPT-based model, we use mean pooling of its hidden states as its output representation, which we found it is better than using only the last token.

\subsection{Linear Probe}
Before fine-tuning each pre-trained models, we first investigate their feature extraction ability by probing their output representations. Probing methods are proposed to determine what information is carried intrinsically by the learned embeddings~\cite{tenney2019you}. We probe the output representation using one single-layer perceptron on top of a ``fixed'' pre-trained language model and only fine-tune that layer for a downstream task with the same hyper-parameters. Table~\ref{tb:probe} shows the probing results of domain classification on MWOZ, intent identification on OOS, and dialogue act prediction on MWOZ. TOD-BERT-jnt achieves the highest performance in this setting, suggesting its representation contains the most useful information. 

\subsection{Intent Recognition}
TOD-BERT outperforms BERT and other strong baselines in one of the largest intent recognition datasets, as shown in Table~\ref{tb:intent}. We evaluate accuracy on all the data, the in-domain intents only, and the out-of-scope intent only. Note that there are two ways to predict out-of-scope intent, one is to treat it as an additional class, and the other is to set a threshold for prediction confidence. Here we report the results of the first setting.
TOD-BERT-jnt achieves the highest in-scope and out-of-scope accuracy.
Besides, we conduct 1-shot and 10-shot experiments by randomly sampling one and ten utterances from each intent class in the training set. 
TOD-BERT-jnt has 13.2\% all-intent accuracy improvement and 16.3\% in-domain accuracy improvement compared to BERT in the 1-shot setting. 

\begin{table*}[!t]
\centering
\resizebox{\linewidth}{!}{
\begin{tabular}{rr|c|c|c|c|c|c}
\hline
\multicolumn{1}{l}{} &  & \multicolumn{2}{c|}{\textbf{MWOZ} (13)} & \multicolumn{2}{c|}{D\textbf{STC2} (19)} & \multicolumn{2}{c}{\textbf{GSIM} (13)} \\ \cline{3-8} 
\multicolumn{1}{l}{} &  & micro-F1 & macro-F1 & micro-F1 & macro-F1 & micro-F1 & macro-F1 \\ \hline
\multicolumn{1}{r|}{\textbf{1\% Data}} & BERT & 84.0\% $\pm$ 0.6\% & 66.7\% $\pm$ 1.7\% & 77.1\% $\pm$ 2.1\% & 25.8\% $\pm$ 0.8\% & 67.3\% $\pm$ 1.4\% & 26.9\% $\pm$ 1.0\% \\
\multicolumn{1}{r|}{} & TOD-BERT-mlm & \textbf{87.5\%} $\pm$ 0.6\% & \textbf{73.3\%} $\pm$ 1.5\% & 79.6\% $\pm$ 1.0\% & 26.4\% $\pm$ 0.5\% & \textbf{82.7\%} $\pm$ 0.7\% & \textbf{35.7\%} $\pm$ 0.3\% \\ 
\multicolumn{1}{r|}{} & TOD-BERT-jnt & 86.9\% $\pm$ 0.2\% & 72.4\% $\pm$ 0.8\% & \textbf{82.9\%} $\pm$ 0.4\% & \textbf{28.0\%} $\pm$ 0.1\% & 78.4\% $\pm$ 3.2\% & 32.9\% $\pm$ 2.1\% \\
\hline

\multicolumn{1}{r|}{\multirow{2}{*}{\textbf{10\% Data}}} & BERT & 89.7\% $\pm$ 0.2\% & 78.4\% $\pm$ 0.3\% & 88.2\% $\pm$ 0.7\% & 34.8\% $\pm$ 1.3\% & 98.4\% $\pm$ 0.3\% & 45.1\% $\pm$ 0.2\% \\
\multicolumn{1}{r|}{} & TOD-BERT-mlm & 90.1\% $\pm$ 0.2\% & 78.9\% $\pm$ 0.1\% & \textbf{91.8\%} $\pm$ 1.7\% & \textbf{39.4\%} $\pm$ 1.7\% & 99.2\% $\pm$ 0.1\% & 45.6\% $\pm$ 0.1\% \\ 
\multicolumn{1}{r|}{} & TOD-BERT-jnt & \textbf{90.2\%} $\pm$ 0.2\% & \textbf{79.6\%} $\pm$ 0.7\% & 90.6\% $\pm$ 3.2\% & 38.8\% $\pm$ 2.2\% & \textbf{99.3\%} $\pm$ 0.1\% & \textbf{45.7\%} $\pm$ 0.0\% \\
\hline

\multicolumn{1}{r|}{\multirow{5}{*}{\textbf{Full Data}}} & MLP & 61.6\% & 45.5\% & 77.6\% & 18.1\% & 89.5\% & 26.1\% \\
\multicolumn{1}{r|}{} & RNN & 90.4\% & 77.3\% & 90.8\% & 29.4\% & 98.4\% & 45.2\% \\
\multicolumn{1}{r|}{} & GPT2 & 90.8\% & 79.8\% & 92.5\% & 39.4\% & 99.1\% & 45.6\% \\
\multicolumn{1}{r|}{} & DialoGPT & 91.2\% & 79.7\% & \textbf{93.8\%} & \textbf{42.1\%} & 99.2\% & 45.6\% \\
\multicolumn{1}{r|}{} & BERT & 91.4\% & 79.7\% & 92.3\% & 40.1\% & 98.7\% & 45.2\% \\ 
\multicolumn{1}{r|}{} & TOD-BERT-mlm & \textbf{91.7\%} & 79.9\% & 90.9\% & 39.9\% & 99.4\% & 45.8\% \\ 
\multicolumn{1}{r|}{} & TOD-BERT-jnt & \textbf{91.7\%} & \textbf{80.6\%} & \textbf{93.8\%} & 41.3\% & \textbf{99.5\%} & \textbf{45.8\%} \\ 
\hline
\end{tabular}
}
\caption{Dialogue act prediction results on three different datasets. The numbers reported are the micro and macro F1 scores, and each dataset has different numbers of dialogue acts.}
\label{tb:da}
\end{table*}

\begin{table}[t]
\centering
\resizebox{\linewidth}{!}{
\begin{tabular}{r|r|c|c}
\hline
 & \textbf{Model} & \textbf{\begin{tabular}[c]{@{}c@{}}Joint\\ Acc\end{tabular}} & \textbf{\begin{tabular}[c]{@{}c@{}}Slot\\ Acc\end{tabular}} \\ \hline

\multirow{2}{*}{\textbf{1\% Data}} 
& BERT & 6.4\% $\pm$ 1.4\% & 84.4\% $\pm$ 1.0\% \\
& TOD-BERT-mlm & \textbf{9.9\%} $\pm$ 0.6\% & \textbf{86.6\%} $\pm$ 0.5\% \\
& TOD-BERT-jnt & 8.0\% $\pm$ 1.0\% & 85.3\% $\pm$ 0.4\% \\
\hline
\multirow{2}{*}{\textbf{5\% Data}} 
& BERT & 19.6\% $\pm$ 0.1\% & 92.0\% $\pm$ 0.5\% \\
& TOD-BERT-mlm & 28.1\% $\pm$ 1.6\% & \textbf{93.9\%} $\pm$ 0.1\% \\ 
& TOD-BERT-jnt & \textbf{28.6\%} $\pm$ 1.4\% & 93.8\% $\pm$ 0.3\% \\
\hline
\multirow{2}{*}{\textbf{10\% Data}} 
& BERT & 32.9\% $\pm$ 0.6\% & 94.7\% $\pm$ 0.1\% \\
& TOD-BERT-mlm & \textbf{39.5\%} $\pm$ 0.7\% & \textbf{95.6\%} $\pm$ 0.1\% \\
& TOD-BERT-jnt & 37.0\% $\pm$ 0.1\% & 95.2\% $\pm$ 0.1\% \\
\hline
\multirow{2}{*}{\textbf{25\% Data}} 
& BERT & 40.8\% $\pm$ 1.0\% & 95.8\% $\pm$ 0.1\% \\
& TOD-BERT-mlm & 44.0\% $\pm$ 0.4\% & \textbf{96.4\%} $\pm$ 0.1\% \\
& TOD-BERT-jnt & \textbf{44.3\%} $\pm$ 0.3\% & 96.3\% $\pm$ 0.2\% \\
\hline
\multirow{7}{*}{\textbf{Full Data}} 
& DSTReader* & 36.4\% & - \\
& HyST* & 38.1\% & - \\
& ZSDST* & 43.4\% & - \\
& TRADE* & 45.6\% & - \\
& GPT2 & 46.2\% & 96.6\% \\
& DialoGPT & 45.2\% & 96.5\% \\
& BERT & 45.6\% & 96.6\% \\ 
& TOD-BERT-mlm & 47.7\% & 96.8\% \\  
& TOD-BERT-jnt & \textbf{48.0\%} & \textbf{96.9\%} \\ 
\hline

\end{tabular}
}
\caption{Dialogue state tracking results on MWOZ 2.1. We report joint goal accuracy and slot accuracy for the full data setting and the simulated few-shot settings.}
\label{tb:dst}
\end{table}

\begin{table*}[t]
\centering
\resizebox{\linewidth}{!}{

\begin{tabular}{rr|c|c|c|c|c|c}
\hline
\multicolumn{1}{r}{} &  & \multicolumn{2}{c|}{\textbf{MWOZ}} & \multicolumn{2}{c|}{\textbf{DSTC2}} & \multicolumn{2}{c}{\textbf{GSIM}} \\ \cline{3-8} 
 &  & 1-to-100 & 3-to-100 & 1-to-100 & 3-to-100 & 1-to-100 & 3-to-100 \\ \hline
 
\multicolumn{1}{c|}{\multirow{2}{*}{\textbf{1\% Data}}} 
& BERT & 7.8\% $\pm$ 2.0\% & 20.5\% $\pm$ 4.4\% & 3.7\% $\pm$ 0.6\% & 9.6\% $\pm$ 1.3\% & 4.0\% $\pm$ 0.4\% & 10.3\% $\pm$ 1.1\% \\
\multicolumn{1}{c|}{} & TOD-BERT-mlm & 13.0\% $\pm$ 1.1\% & 34.6\% $\pm$ 0.4\% & 12.5\% $\pm$ 6.7\% & 24.9\% $\pm$ 10.7\% & 7.2\% $\pm$ 4.0\% & 15.4\% $\pm$ 8.0\% \\ 
\multicolumn{1}{c|}{} & TOD-BERT-jnt & - & - & \textbf{37.5\%} $\pm$ 0.6\% & \textbf{55.9\%} $\pm$ 0.4\% & \textbf{12.5\%} $\pm$ 0.9\% & \textbf{26.8\%} $\pm$ 0.8\% \\
\hline

\multicolumn{1}{c|}{\multirow{2}{*}{\textbf{10\% Data}}} 
& BERT & 20.9\% $\pm$ 2.6\% & 45.4\% $\pm$ 3.8\% & 8.9\% $\pm$ 2.3\% & 21.4\% $\pm$ 3.1\% & 9.8\% $\pm$ 0.1\% & 24.4\% $\pm$ 1.2\% \\
\multicolumn{1}{c|}{} & TOD-BERT-mlm & 22.3\% $\pm$ 3.2\% & 48.7\% $\pm$ 4.0\% & 19.0\% $\pm$ 16.3\% & 33.8\% $\pm$ 20.4\% & 11.2\% $\pm$ 2.5\% & 26.0\% $\pm$ 2.7\% \\ 
\multicolumn{1}{c|}{} & TOD-BERT-jnt & - & - & \textbf{49.7\%} $\pm$ 0.3\% & \textbf{66.6\%} $\pm$ 0.1\% & \textbf{23.0\%} $\pm$ 1.0\% & \textbf{42.6\%} $\pm$ 1.0\% \\
\hline

\multicolumn{1}{r|}{\multirow{5}{*}{\textbf{Full Data}}} & GPT2 & 47.5\% & 75.4\% & 53.7\% & 69.2\% & 39.1\% & 60.5\% \\
\multicolumn{1}{l|}{} & DialoGPT & 35.7\% & 64.1\% & 39.8\% & 57.1\% & 16.5\% & 39.5\% \\
\multicolumn{1}{l|}{} & BERT & 47.5\% & 75.5\% & 46.6\% & 62.1\% & 13.4\% & 32.9\% \\ 
\multicolumn{1}{l|}{} & TOD-BERT-mlm & 48.1\% & 74.3\% & 50.0\% & 65.1\% & 36.5\% & 60.1\% \\ 
\multicolumn{1}{l|}{} & TOD-BERT-jnt & \textbf{65.8\%} & \textbf{87.0\%} & \textbf{56.8\%} & \textbf{70.6\%} & \textbf{41.0\%} & \textbf{65.4\%} \\
\hline
\end{tabular}
}
\caption{Response selection evaluation results on three corpora for 1\%, 10\% and full data setting. We report 1-to-100 and 3-to-100 accuracy, which is similar to recall\@1 and recall@3 given 100 candidates.}
\label{tb:rs}
\end{table*}

\subsection{Dialogue State Tracking}
Two evaluation metrics are commonly used in dialogue state tracking task: joint goal accuracy and slot accuracy. The joint goal accuracy compares the predicted dialogue states to the ground truth at each dialogue turn. The ground truth includes slot values for all the possible (domain, slot) pairs. The output is considered as a correct prediction if and only if all the predicted values exactly match its ground truth values. On the other hand, the slot accuracy individually compares each (domain, slot, value) triplet to its ground truth label. 

In Table~\ref{tb:dst}, we compare BERT to TOD-BERT-jnt on the MWOZ 2.1 dataset and find the latter has 2.4\% joint goal accuracy improvement. Since the original ontology provided by \citet{budzianowski2018multiwoz} is not complete (some labeled values are not included in the ontology), we create a new ontology of all the possible annotated values. We also list several well-known dialogue state trackers as reference, including DSTReader \cite{gao2019dialog}, HyST \cite{goel2019hyst}, TRADE \cite{wu-etal-2019-transferable}, and ZSDST \cite{rastogi2019towards}.
We also report the few-shot experiments using 1\%, 5\%, 10\%, and 25\% data. Note that 1\% of data has around 84 dialogues. TOD-BERT outperforms BERT in all the setting, which further show the strength of task-oriented dialogue pre-training. 


\subsection{Dialogue Act Prediction}
We conduct experiments on three different datasets and report micro-F1 and macro-F1 scores for the dialogue act prediction task, a multi-label classification problem. For the MWOZ dataset, we remove the domain information from the original system dialogue act labels. For example, the ``taxi-inform'' will be simplified to ``inform''. This process reduces the number of possible dialogue acts from 31 to 13. For DSTC2 and GSIM corpora, we follow \citet{paul2019towards} to apply universal dialogue act mapping that maps the original dialogue act labels to a general dialogue act format, resulting in 19 and 13 system dialogue acts in DSTC2 and GSIM, respectively. We run two other baselines, MLP and RNN, to further show the strengths of BERT-based models. The MLP model simply takes bag-of-word embeddings to make dialogue act prediction, and the RNN model is a bi-directional GRU network. 

In Table~\ref{tb:da}, one can observe that in full data scenario, TOD-BERT consistently works better than BERT and other baselines, no matter which datasets or which evaluation metrics. In the few-shot experiments, TOD-BERT-mlm outperforms BERT by 3.5\% micro-F1 and 6.6\% macro-F1 on MWOZ corpus in the 1\% data scenario. We also found that 10\% of training data can achieve good performance that is close to full data training. 

\begin{figure}[t!]
    \centering
    \subfloat[BERT]{
        \includegraphics[width=0.485\linewidth]{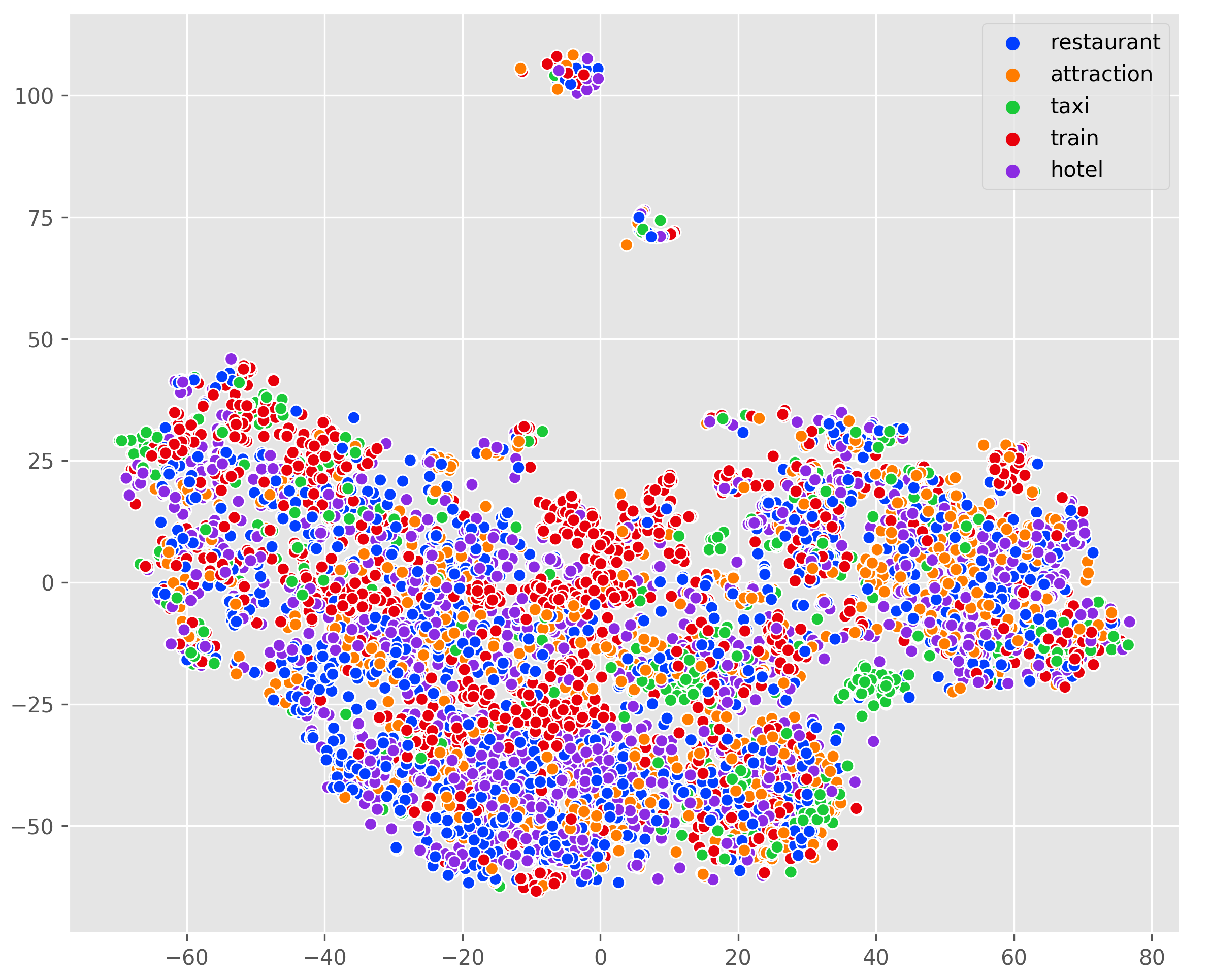}} 
    \hfill
    \subfloat[BERT]{
        \includegraphics[width=0.485\linewidth]{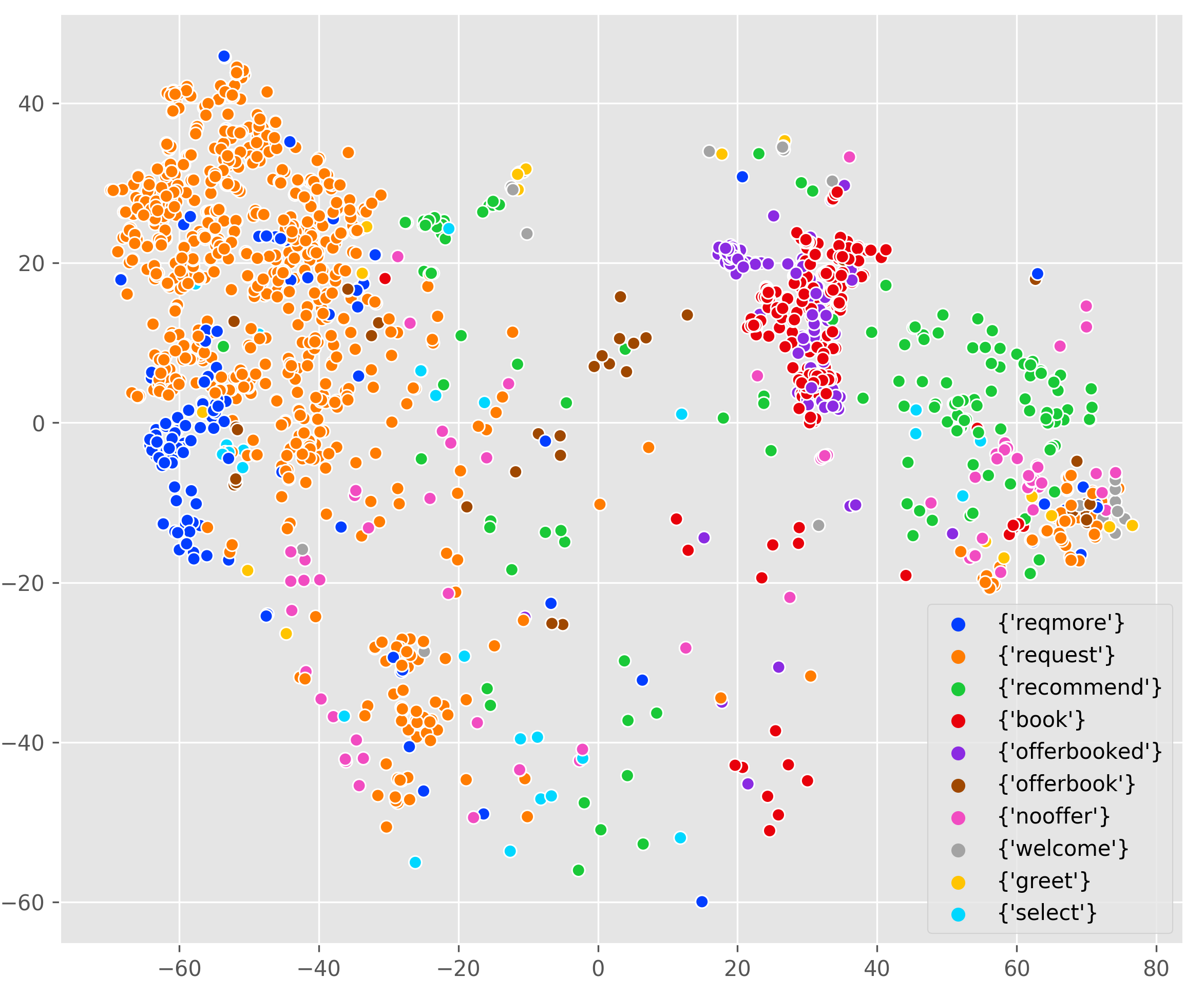}}
    \\
    \subfloat[TOD-BERT-mlm]{
        \includegraphics[width=0.485\linewidth]{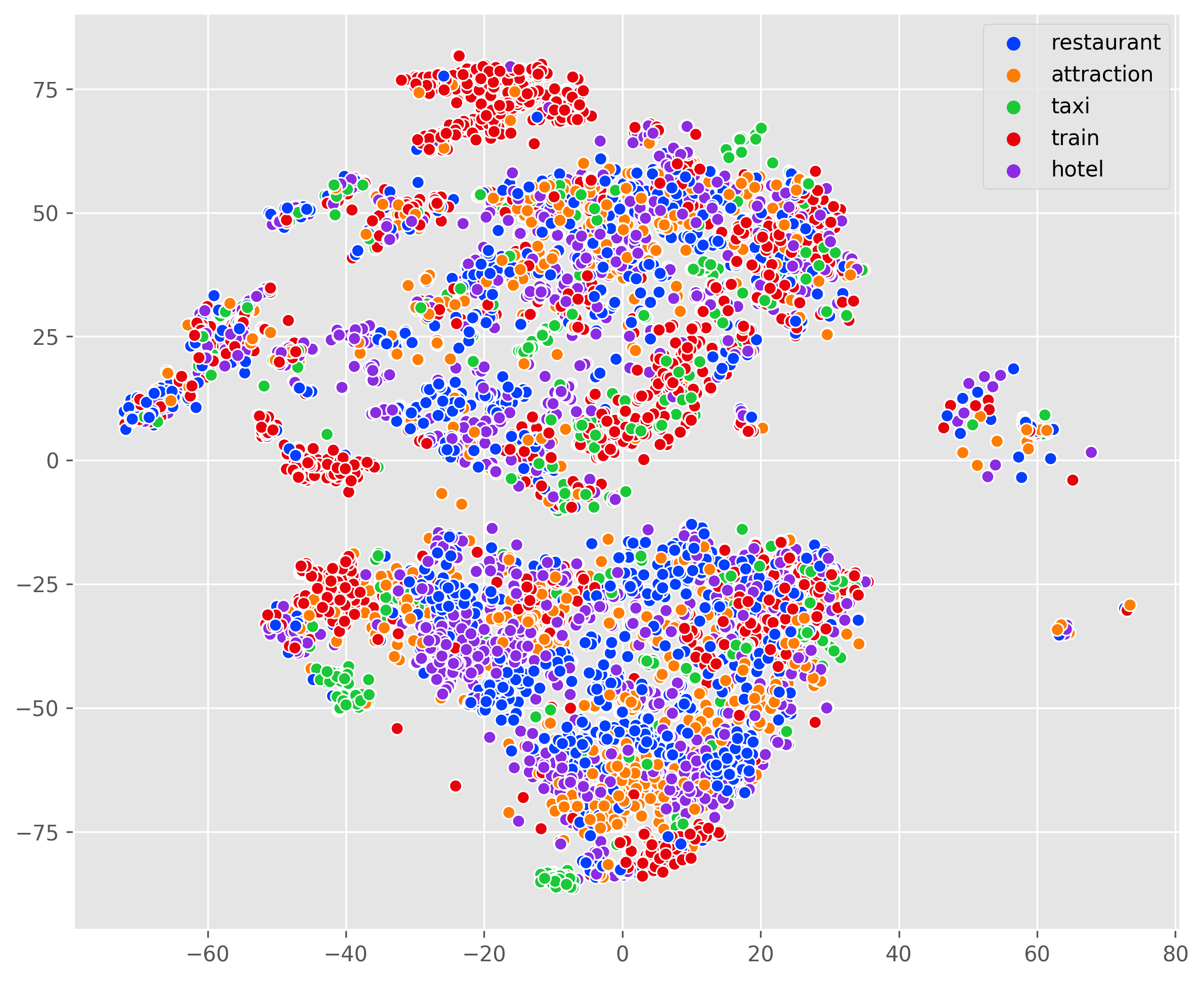}} 
    \hfill
    \subfloat[TOD-BERT-mlm]{
        \includegraphics[width=0.485\linewidth]{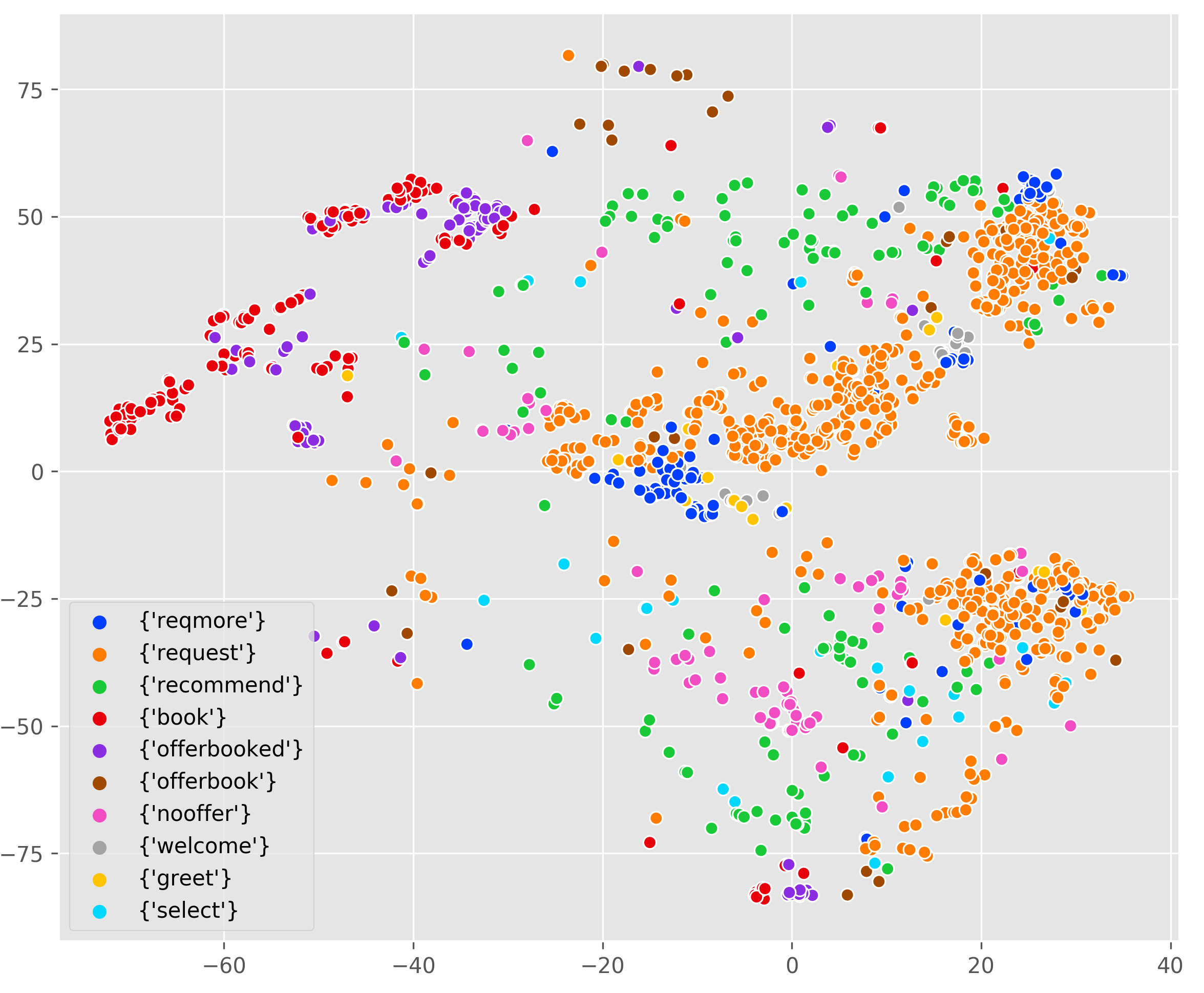}}
    \\
    \subfloat[TOD-BERT-jnt]{
        \includegraphics[width=0.485\linewidth]{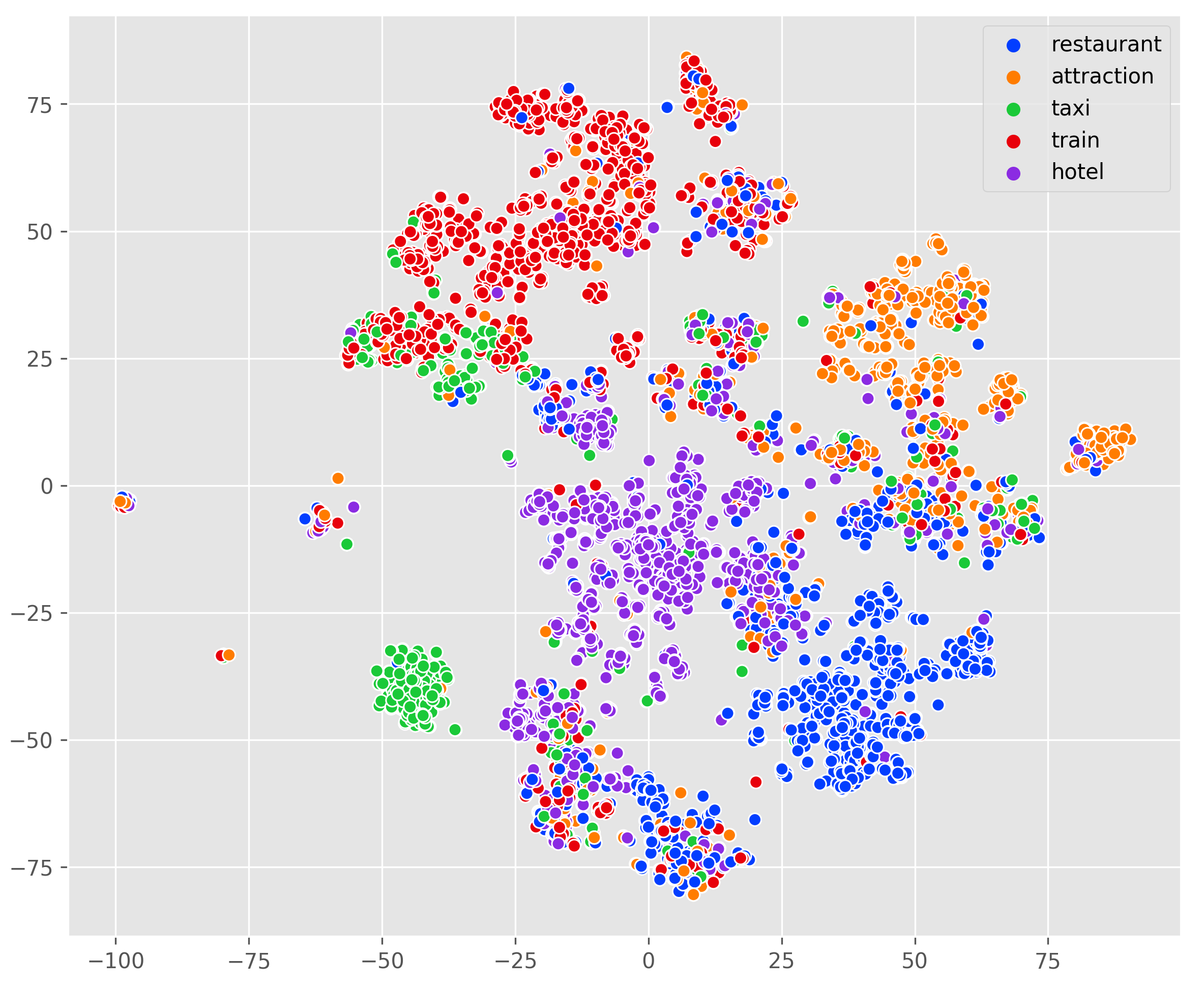}} 
    \hfill
    \subfloat[TOD-BERT-jnt]{
        \includegraphics[width=0.485\linewidth]{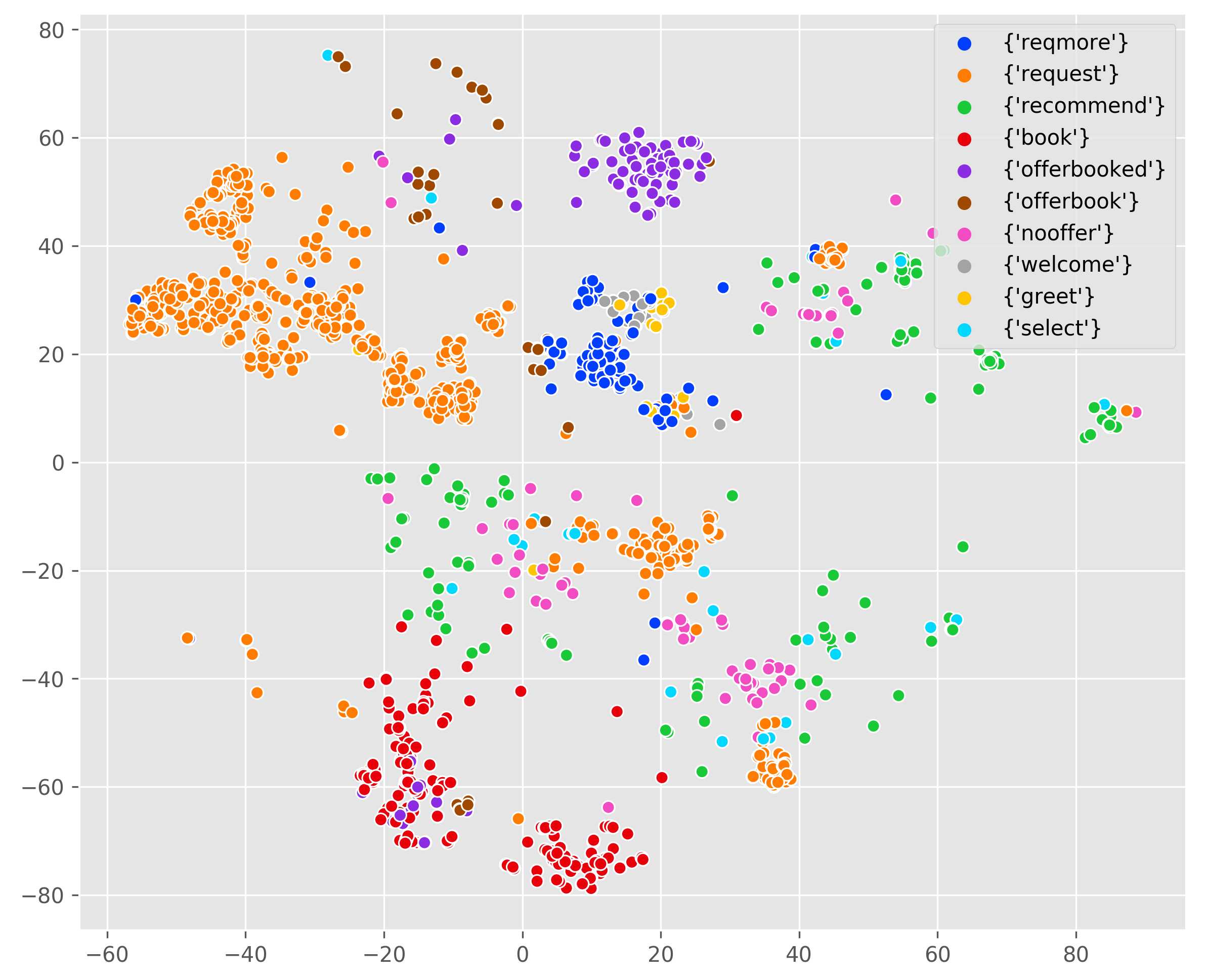}}
    
    \caption{The tSNE visualization of BERT, TOD-BERT-mlm and TOD-BERT-jnt representations of system responses in the MWOZ test set. Different colors in the left-hand column mean different domains, and in the right-hand column represent different dialogue acts.}
    \label{fig:tsne}
\end{figure}

\subsection{Response Selection}
To evaluate response selection in task-oriented dialogues, we follow the k-to-100 accuracy, which is becoming a research community standard \cite{yang-etal-2018-learning, henderson2019convert}. 
The k-of-100 metric is computed using a random batch of 100 examples so that responses from other examples in the same batch can be used as random negative candidates. 
This allows us to be compute the metric across many examples in batches efficiently. 
While it is not guaranteed that the random negatives will indeed be ``true'' negatives, the 1-of-100 metric still provides a useful evaluation signal. During inference, we run five different random seeds to sample batches and report the average results.

In Table~\ref{tb:rs}, we conduct response selection experiments on three datasets, MWOZ, DSTC2, and GSIM. TOD-BERT-jnt achieves 65.8\% 1-to-100 accuracy and 87.0\% 3-to-100 accuracy on MWOZ, which surpasses BERT by 18.3\% and 11.5\%, respectively. 
The similar results are also consistently observed in DSTC2 and GSIM datasets, and the advantage of the TOD-BERT-jnt is more evident in the few-shot scenario. 
We do not report TOD-BERT-jnt for MWOZ few-shot setting because it is not fair to compare them with others as the full MWOZ training set is used for response contrastive learning during pre-training stage.
The response selection results are sensitive to the training batch size since the larger the batch size the harder the prediction. In our experiments, we set batch size equals to 25 for all the models.

\section{Visualization}
\label{sec:Discussion}
In Figure~\ref{fig:tsne}, we visualize the embeddings of BERT, TOD-BERT-mlm, and TOD-BERT-jnt given the same input from the MWOZ test set. Each sample point is a system response representation, which is passed through a pre-trained model and reduced its high-dimension features to a two-dimension point using the t-distributed stochastic neighbor embedding (tSNE) for dimension reduction. Since we know the true domain and dialogue act labels for each utterance, we use different colors to represent different domains and dialogue acts. As one can observe, TOD-BERT-jnt has more clear group boundaries than TOD-BERT-mlm, and two of them are better than BERT.

To analyze the results quantitatively, we run K-means, a common unsupervised clustering algorithms, on top of the output embeddings of BERT and TOD-BERT. We set K for K-means equal to 10 and 20. After the clustering, we can assign each utterance in the MWOZ test set to a predicted class. We then compute the normalized mutual information (NMI) between the clustering result and the actual domain label for each utterance. Here is what we observe: TOD-BERT consistently achieves higher NMI scores than BERT. For K=10, TOD-BERT has a 0.143 NMI score, and BERT only has 0.094. For K=20, TOD-BERT achieves a 0.213 NMI score, while BERT has 0.109.


\section{Conclusion}
\label{sec:Conclusion}
We propose task-oriented dialogue BERT (TOD-BERT) trained on nine human-human and multi-turn task-oriented datasets across over 60 domains. TOD-BERT outperforms BERT on four dialogue downstream tasks, including intention classification, dialogue state tracking, dialogue act prediction, and response selection. It also has a clear advantage in the few-shot experiments when only limited labeled data is available. TOD-BERT is easy-to-deploy and will be open-sourced, allowing the NLP research community to apply or fine-tune any task-oriented conversational problem. 




\bibliography{anthology,emnlp2020}
\bibliographystyle{acl_natbib}

\clearpage
\newpage

\appendix
\input{Appendix}






\end{document}

%% file: Appendix.tex
\section{Appendices}

\begin{figure}[h!]
    \centering
    \subfloat[BERT]{
        \includegraphics[width=0.8\linewidth]{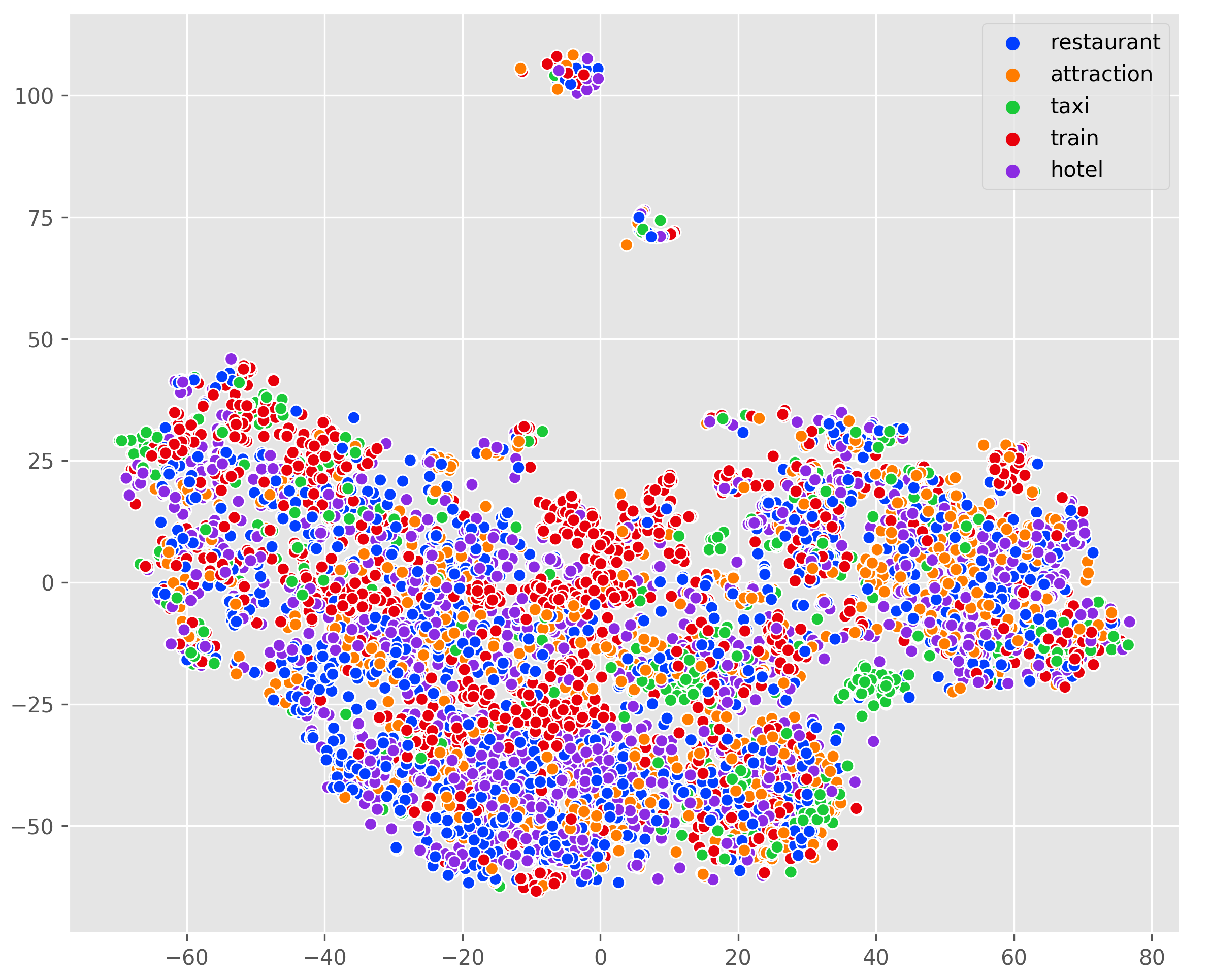}} 
    \hfill
    \subfloat[TOD-BERT-mlm]{
        \includegraphics[width=0.8\linewidth]{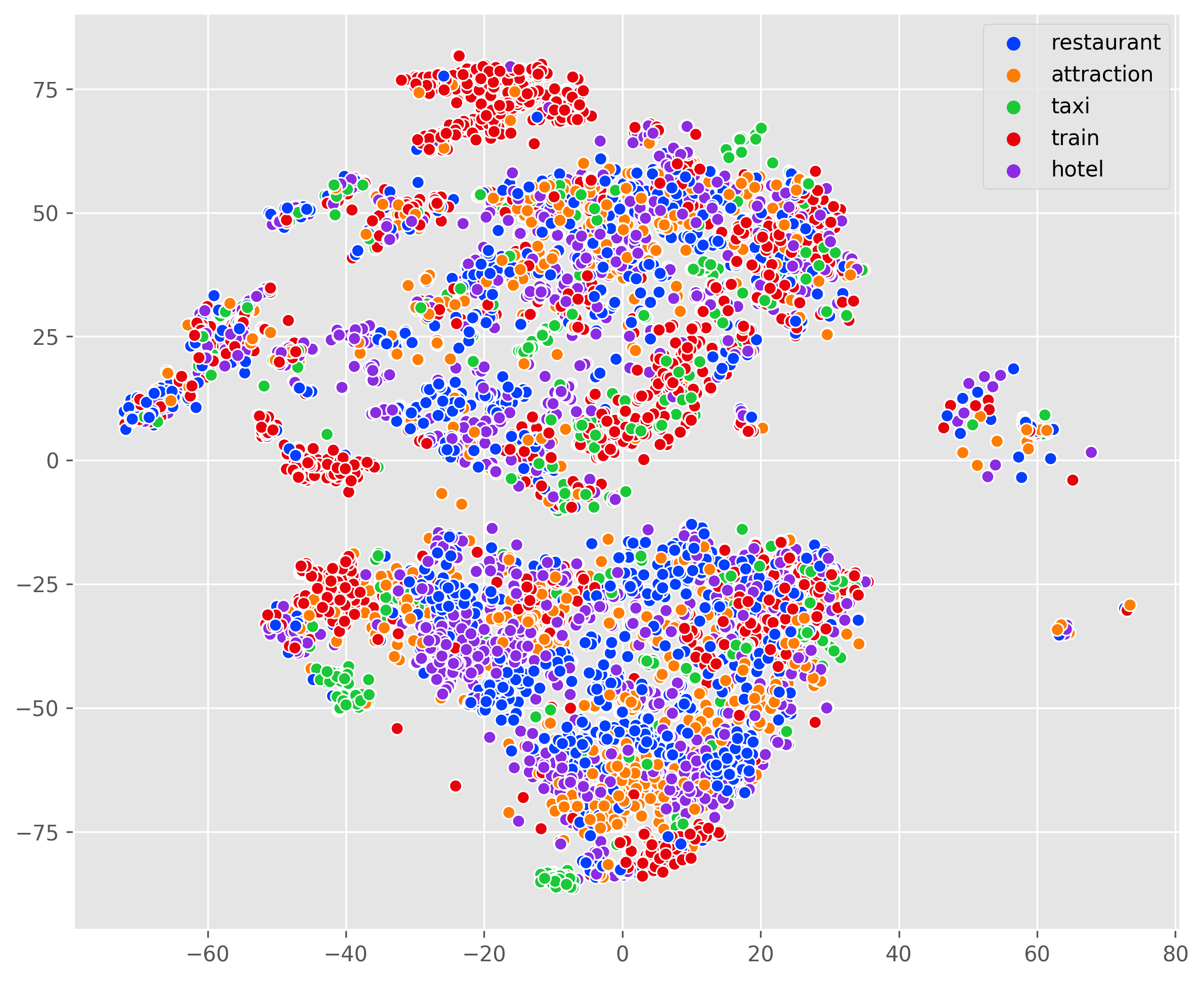}}
    \hfill
    \subfloat[TOD-BERT-jnt]{
        \includegraphics[width=0.8\linewidth]{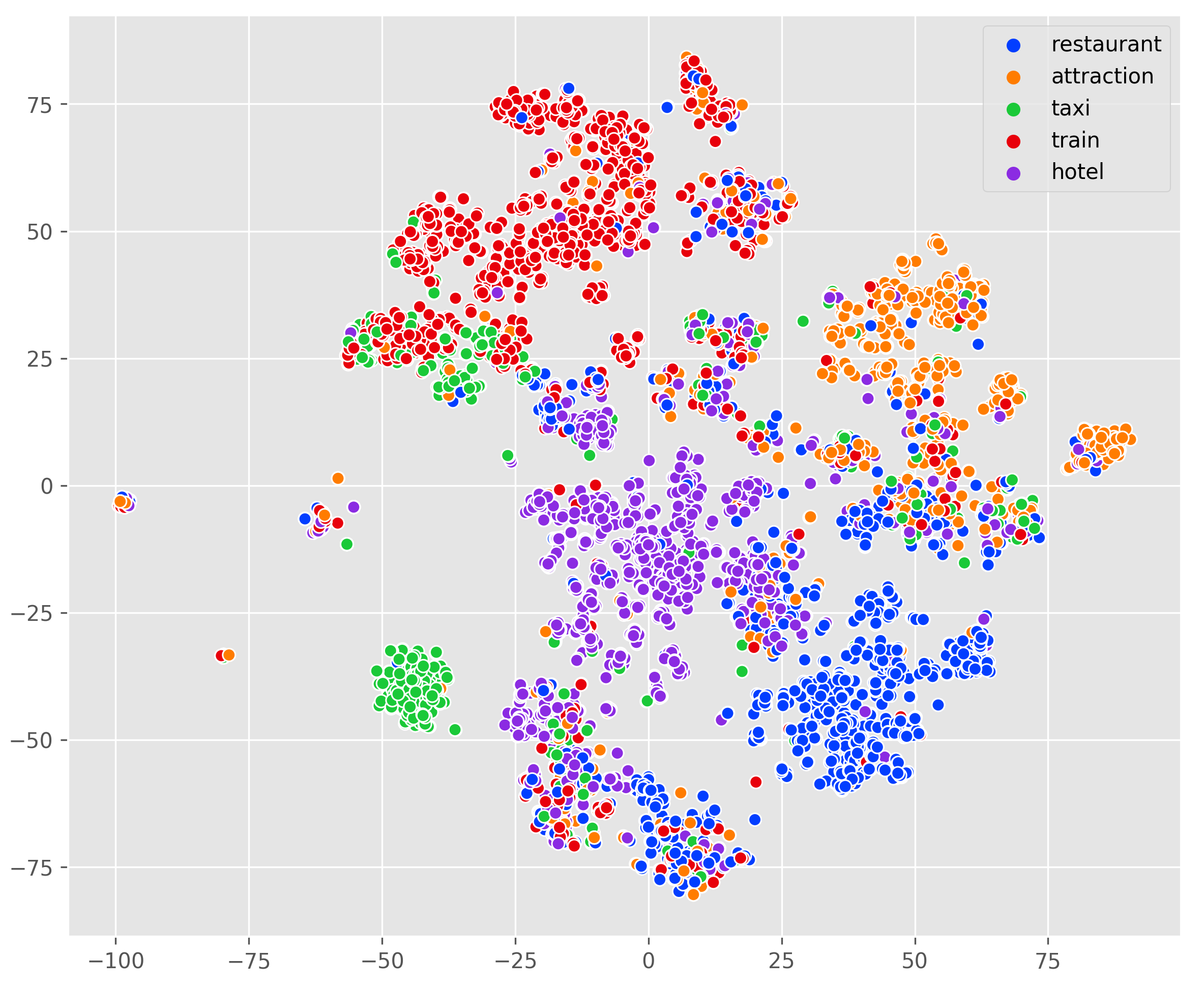}}
    
    \caption{The tSNE visualization of BERT and TOD-BERT representations of system responses in MWOZ test set. Different colors mean different domains.}
    \label{fig:tsne1}
\end{figure}

\begin{figure}[t!]
    \centering
    \subfloat[BERT]{
        \includegraphics[width=0.8\linewidth]{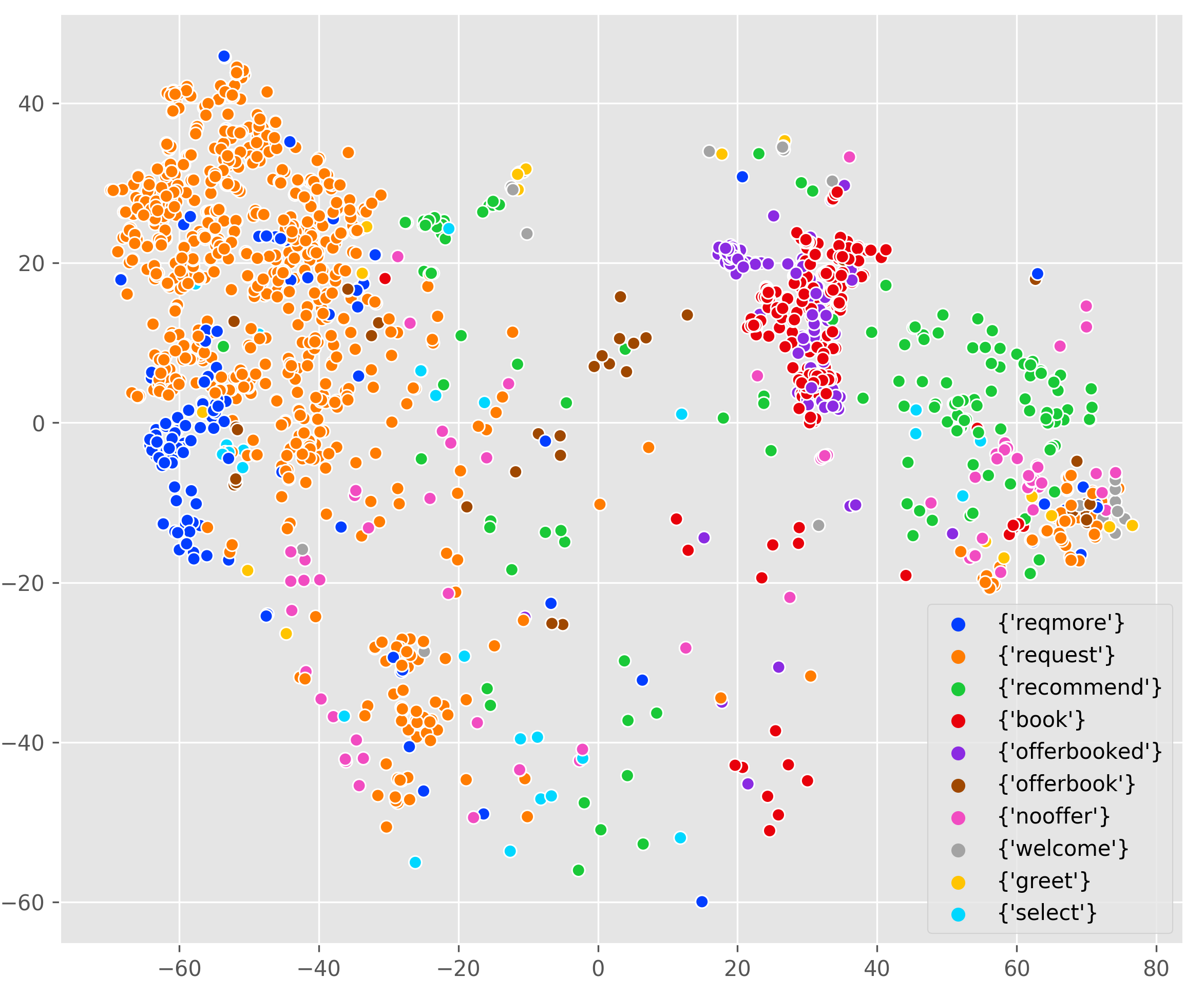}} 
    \hfill
    \subfloat[TOD-BERT-mlm]{
        \includegraphics[width=0.8\linewidth]{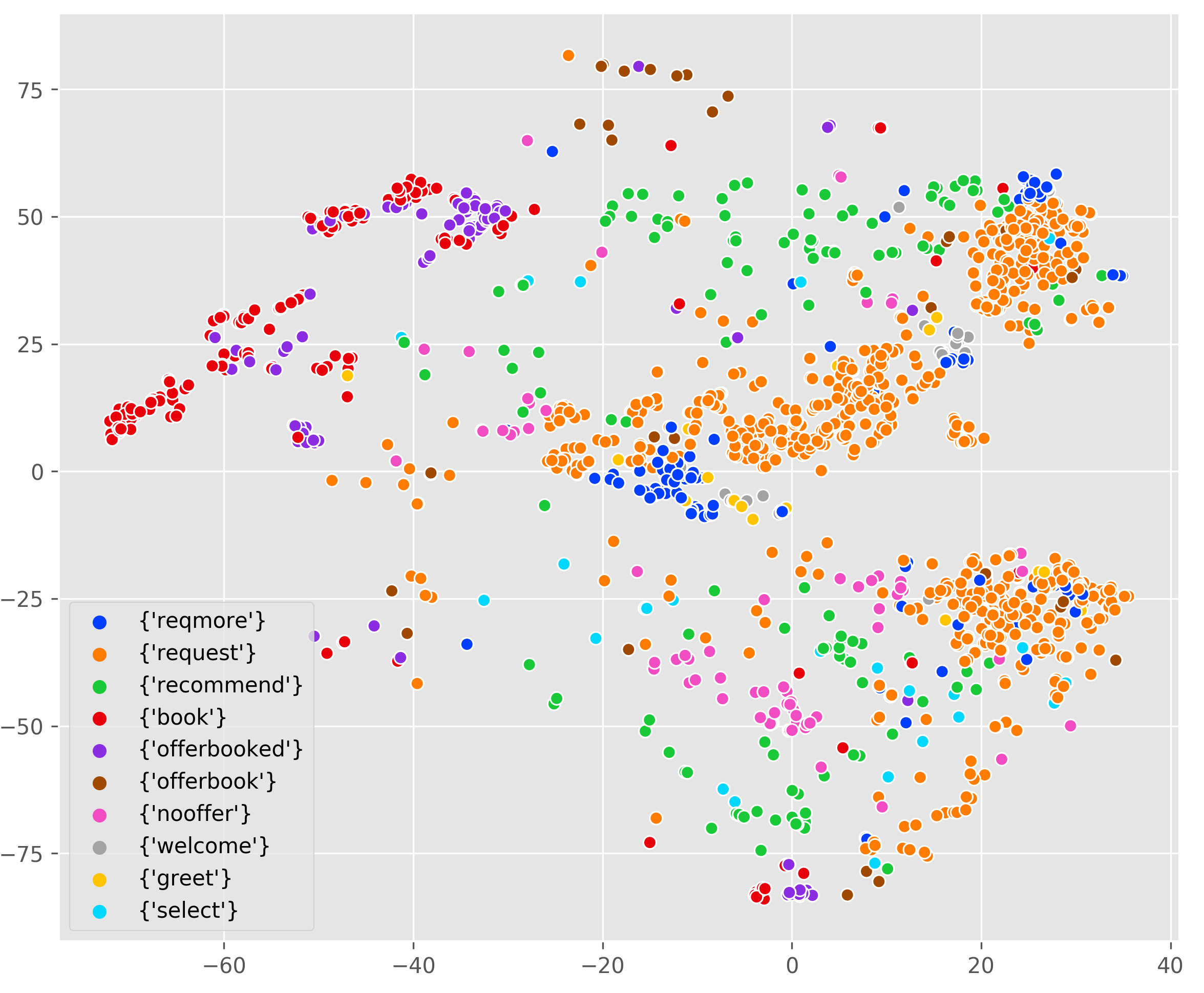}}
    \hfill
    \subfloat[TOD-BERT-jnt]{
        \includegraphics[width=0.8\linewidth]{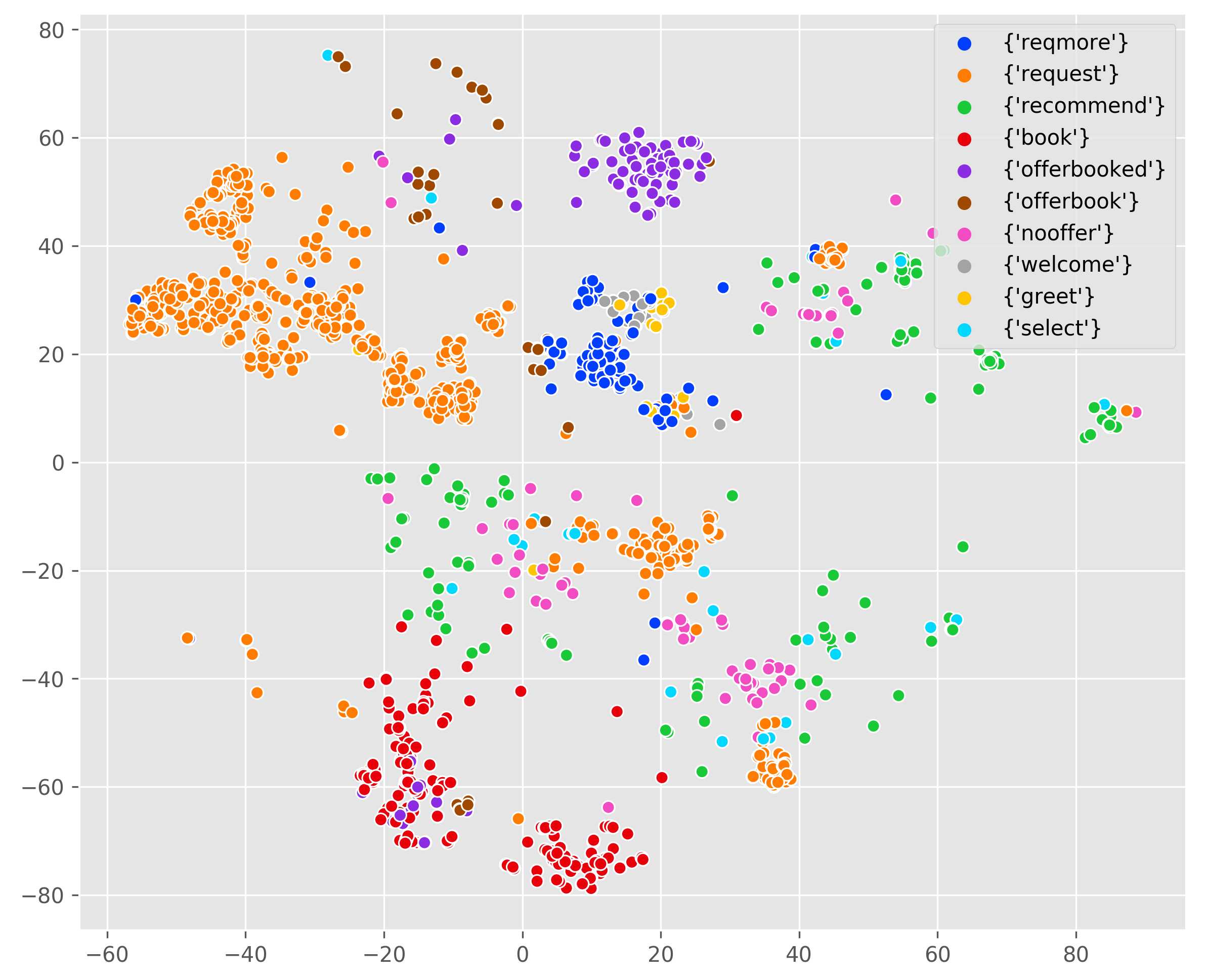}}
    
    \caption{The tSNE visualization of BERT and TOD-BERT representations of system responses in MWOZ test set. Different colors mean different dialogue acts.}
    \label{fig:tsne2}
\end{figure}

\begin{figure}[h!]
    \centering
    \subfloat[BERT]{
        \includegraphics[width=0.8\linewidth]{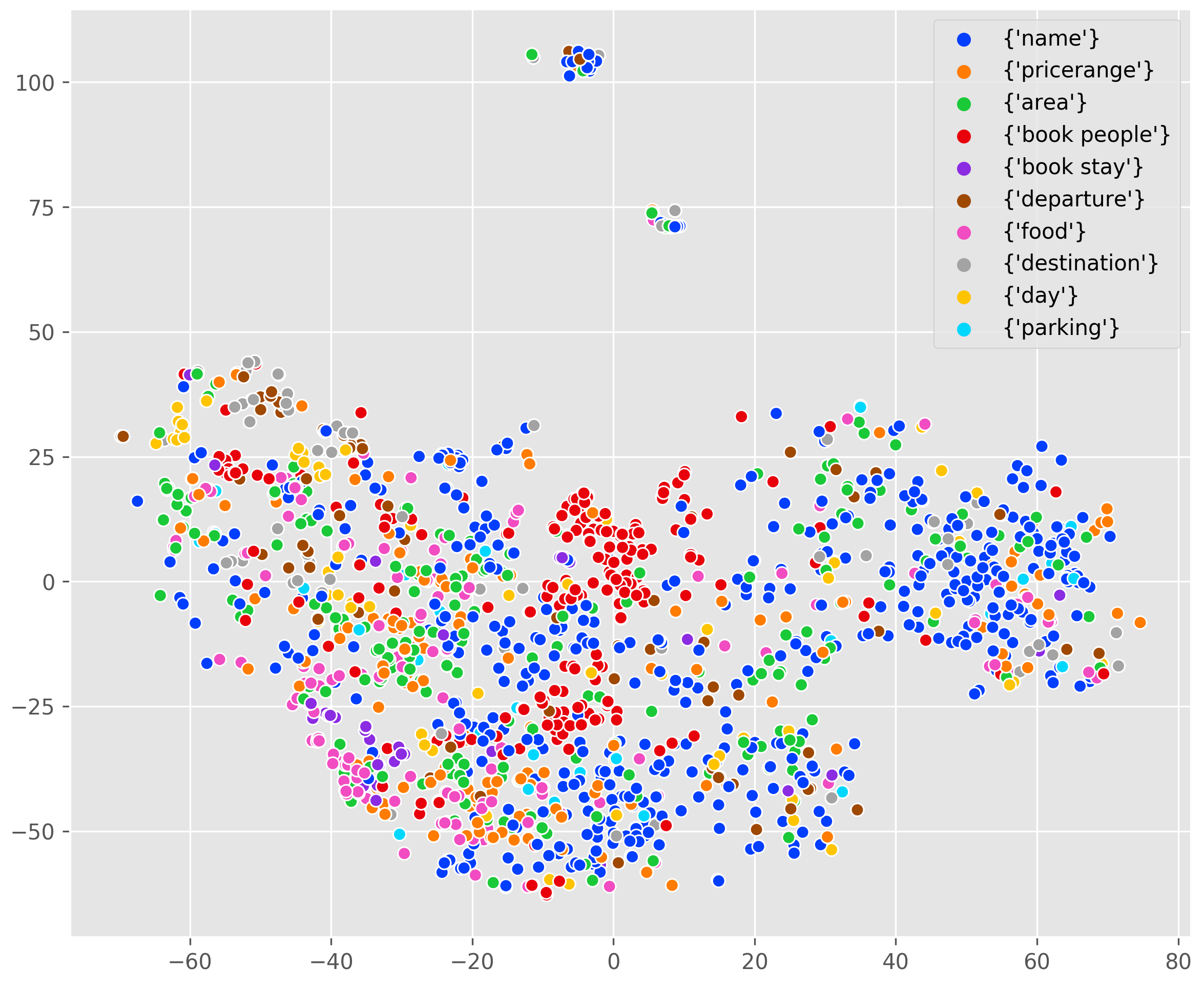}} 
    \hfill
    \subfloat[TOD-BERT-mlm]{
        \includegraphics[width=0.8\linewidth]{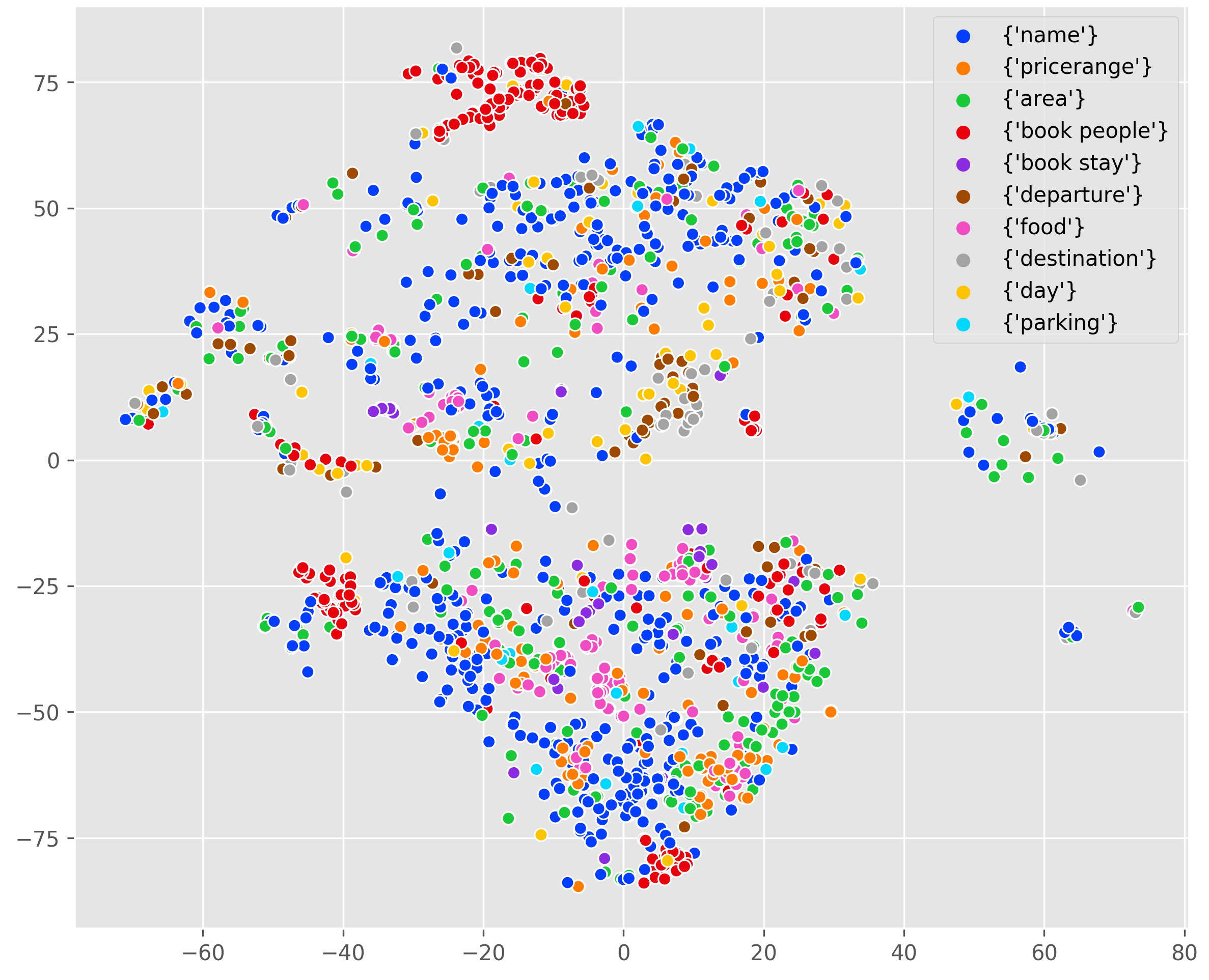}}
    \hfill
    \subfloat[TOD-BERT-jnt]{
        \includegraphics[width=0.8\linewidth]{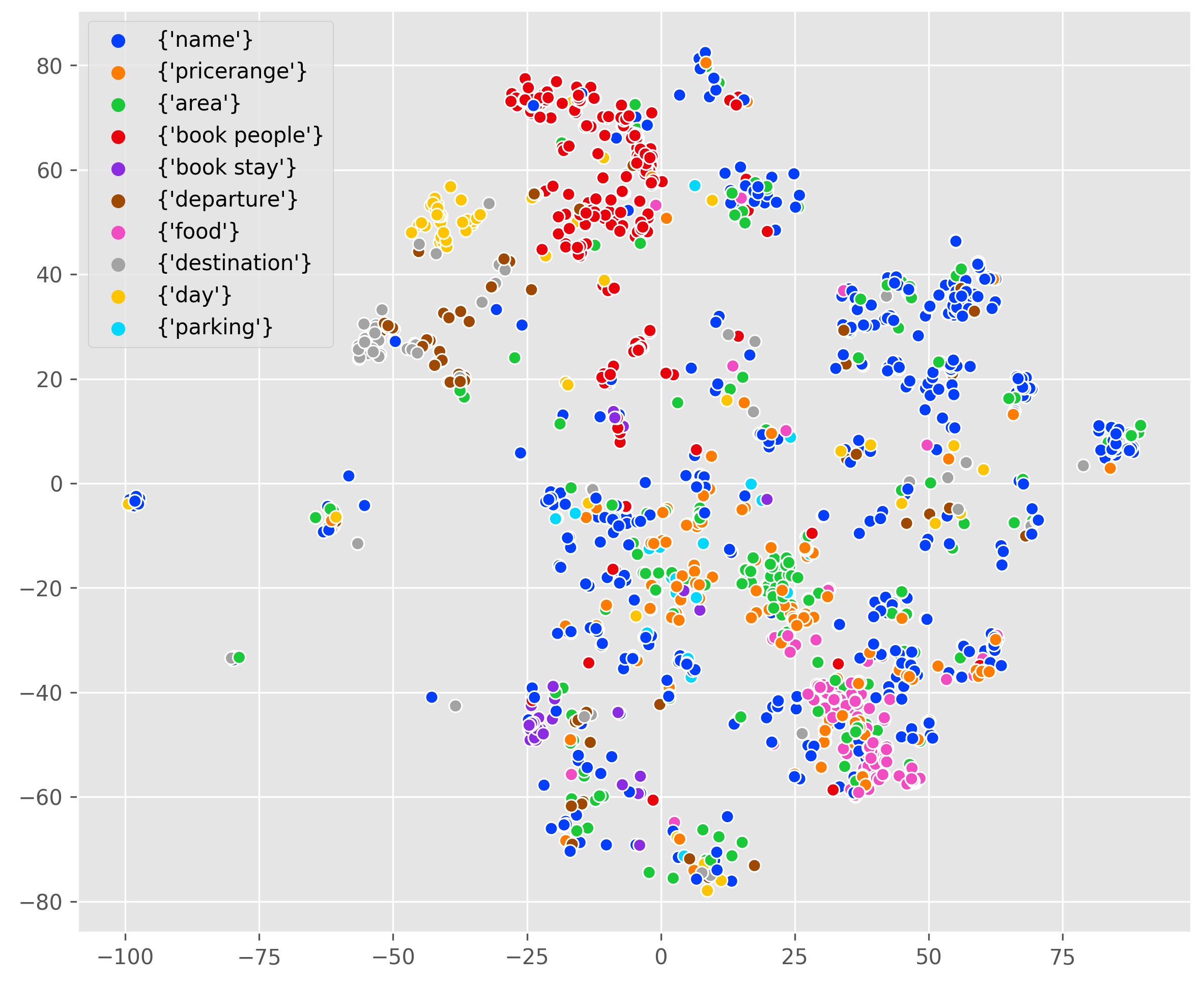}}
    
    \caption{The tSNE visualization of BERT and TOD-BERT representations of system responses in MWOZ test set. Different colors mean different dialogue slots.}
    \label{fig:tsne3}
\end{figure}